\documentclass[11pt]{article}

\usepackage[final]{acl}

\usepackage{times}
\usepackage{latexsym}

\usepackage[T1]{fontenc}

\usepackage[utf8]{inputenc}

\usepackage{microtype}

\usepackage{inconsolata}

\usepackage{graphicx}
\usepackage{xcolor}         
\usepackage{amsmath}
\usepackage{wrapfig}
\usepackage{booktabs}
\usepackage{tabularx}
\usepackage{subcaption}
\usepackage{amsfonts}
\usepackage{multirow}
\usepackage{algorithm}
\usepackage{algpseudocode}

\title{Counteraction-Aware Multi-Teacher On-Policy Distillation for General Capability Recovery with Domain Preservation}

\author{
  \textbf{Tianlei Chen}\textsuperscript{1,*\textdaggerdbl},
  \textbf{Jiao Ou}\textsuperscript{1,*\dag},
  \textbf{Ziyuan Liu}\textsuperscript{1},
  \textbf{Ruiming Tang}\textsuperscript{1,\dag},
  \textbf{Jian Liang}\textsuperscript{1},
  \textbf{Han Li}\textsuperscript{1} \\
  \textsuperscript{1}Kuaishou Technology, Beijing, China \\
  \texttt{tianlei\_chen@163.com, ojiao1111@gmail.com, tangruiming@kuaishou.com}
}

\begin{document}
\maketitle
\begingroup
\renewcommand{\thefootnote}{\fnsymbol{footnote}}
\footnotetext[1]{Equal contribution.}
\footnotetext[2]{Corresponding authors.}
\footnotetext[3]{Work completed during an internship at Kuaishou Technology.}
\endgroup
\begin{abstract}
Domain specialization can improve LLM behavior, but often weakens the general capabilities inherited from the original model. Recent Multi-Teacher On-Policy Distillation (MOPD) pipelines recover model capabilities by supervising student-generated trajectories with teacher feedback, but typically assume teacher-aligned prompt coverage, requiring prompts to match the teachers' training distributions. This assumption is difficult to satisfy when the general teacher is an open-source model whose post-training data are unknown. Instead of attempting to reconstruct this hidden distribution, we study general capability recovery with readily available proxy general prompts. We identify two failure modes of vanilla MOPD in this incomplete-coverage situation: recovery-preservation counteraction from mixing conflicting recovery and preservation gradients, and weak-signal flattening from uniformly averaging samples with unequal correction demand. We propose \textbf{Counteraction-Aware Multi-Teacher On-Policy Distillation} (\textbf{CaMOPD}), which addresses these issues with decoupled alternating training and gap-based sample selection. CaMOPD allocates dedicated updates to general recovery, periodically performs domain-preservation updates, and selects samples with larger averaged token-level teacher-student log-probability gaps to concentrate correction signals. Across role-play dialogue and medical reasoning QA scenarios, CaMOPD outperforms all baselines in general capability recovery while maintaining domain-specific behavior. Gradient coherence analyses further support the intended effect of CaMOPD in producing more coherent correction signals.
\end{abstract}


\section{Introduction}

The remarkable success of large language models (LLMs) has accelerated their deployment across vertical domains, from role-play~\citep{coser} to clinical decision support~\citep{baichuan_m3}. However, domain specialization faces a fundamental paradox: the process of specialization itself often weakens the general capabilities that made these models valuable in the first place, creating a need for capability preservation~\citep{catastrophic_interference, finetuning_distortion}. Classical mitigation strategies mix general-domain data or use replay mechanisms to reduce forgetting~\citep{gem, experience_replay_cl, sft_data_composition, self_synthesized_rehearsal}. However, these approaches heavily rely on massive amounts of high-quality original data and introduce challenges in optimizing training schedules.

Recently, multi-teacher on-policy distillation (MOPD) has attracted attention as a way to recover diverse capabilities into a single student model by supervising student-generated trajectories with token-level teacher feedback~\citep{nemotron_cascade2, glm5}. However, existing MOPD pipelines often assume teacher-aligned prompt coverage: the prompts used during MOPD should sufficiently match the data distribution that produced the teachers~\citep{mimo_mopd, baichuan_m3}. This premise places a heavy burden on data collection and can be expensive, time-consuming, or impossible when teacher's original post-training data are unavailable.

\begin{figure}[t]
    \centering
    \begin{subfigure}[t]{\linewidth}
        \centering
        \includegraphics[width=\linewidth]{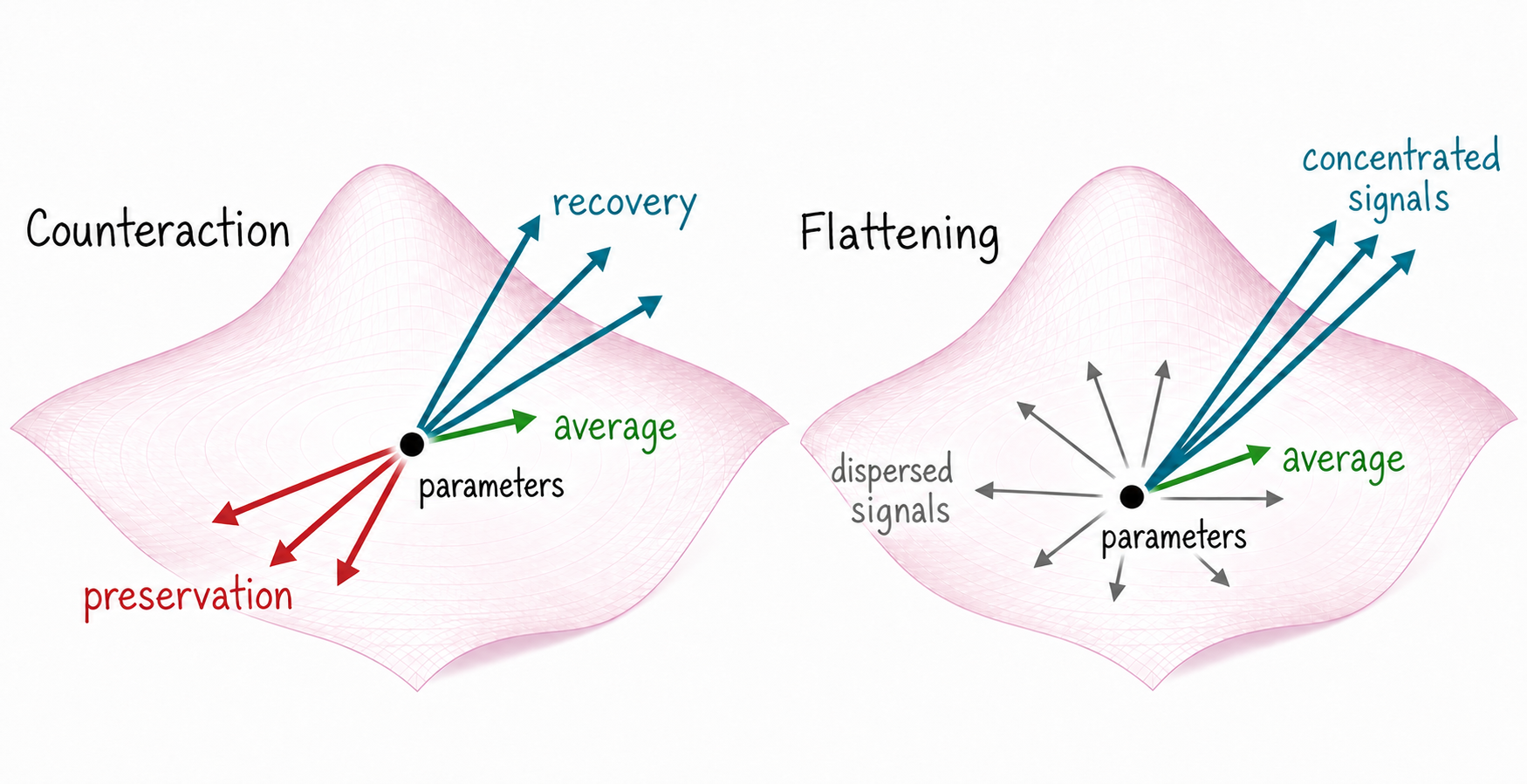}
    \end{subfigure}
    \begin{subfigure}[t]{0.49\linewidth}
        \centering
        \includegraphics[width=\linewidth]{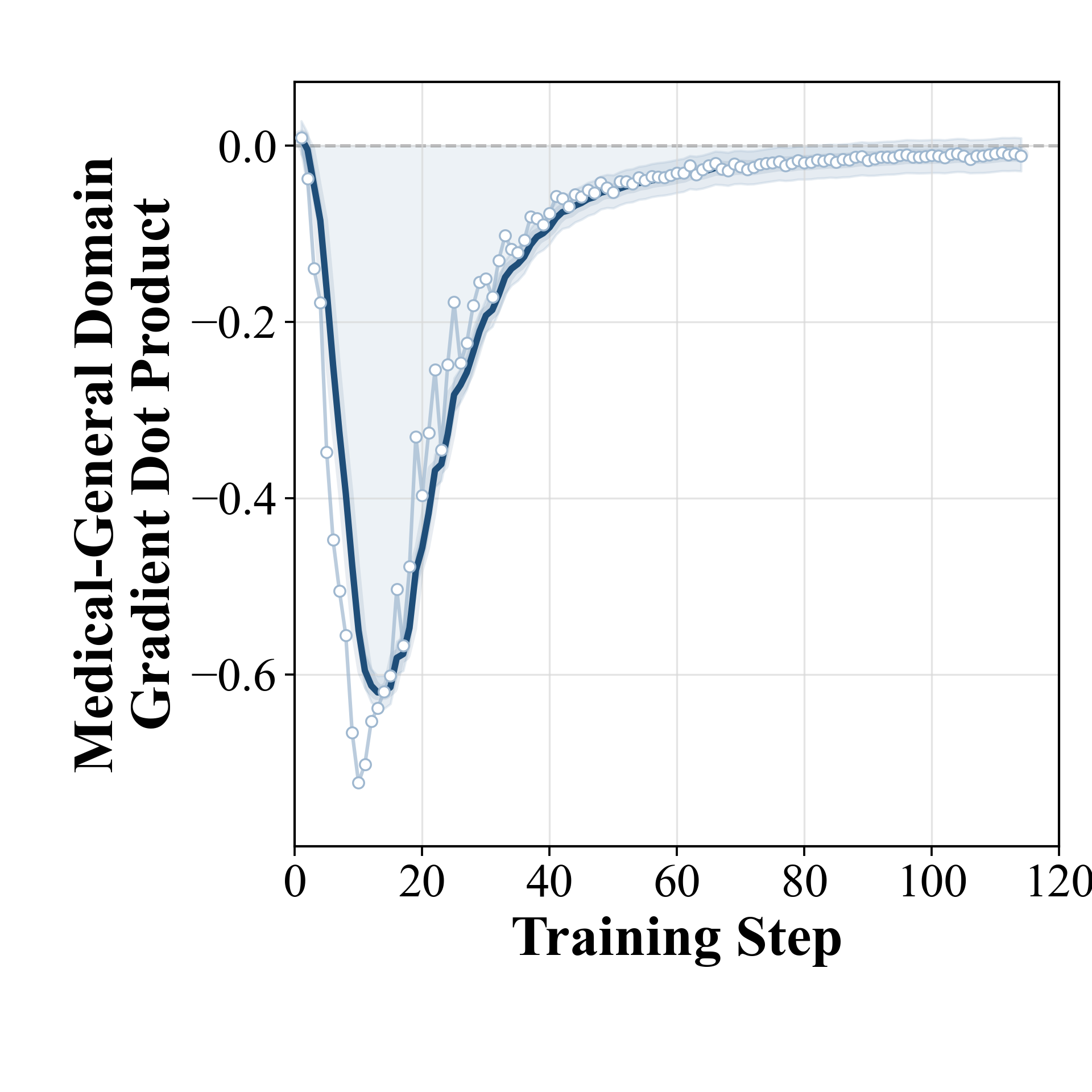}
    \end{subfigure}
    \begin{subfigure}[t]{0.49\linewidth}
        \centering
        \includegraphics[width=\linewidth]{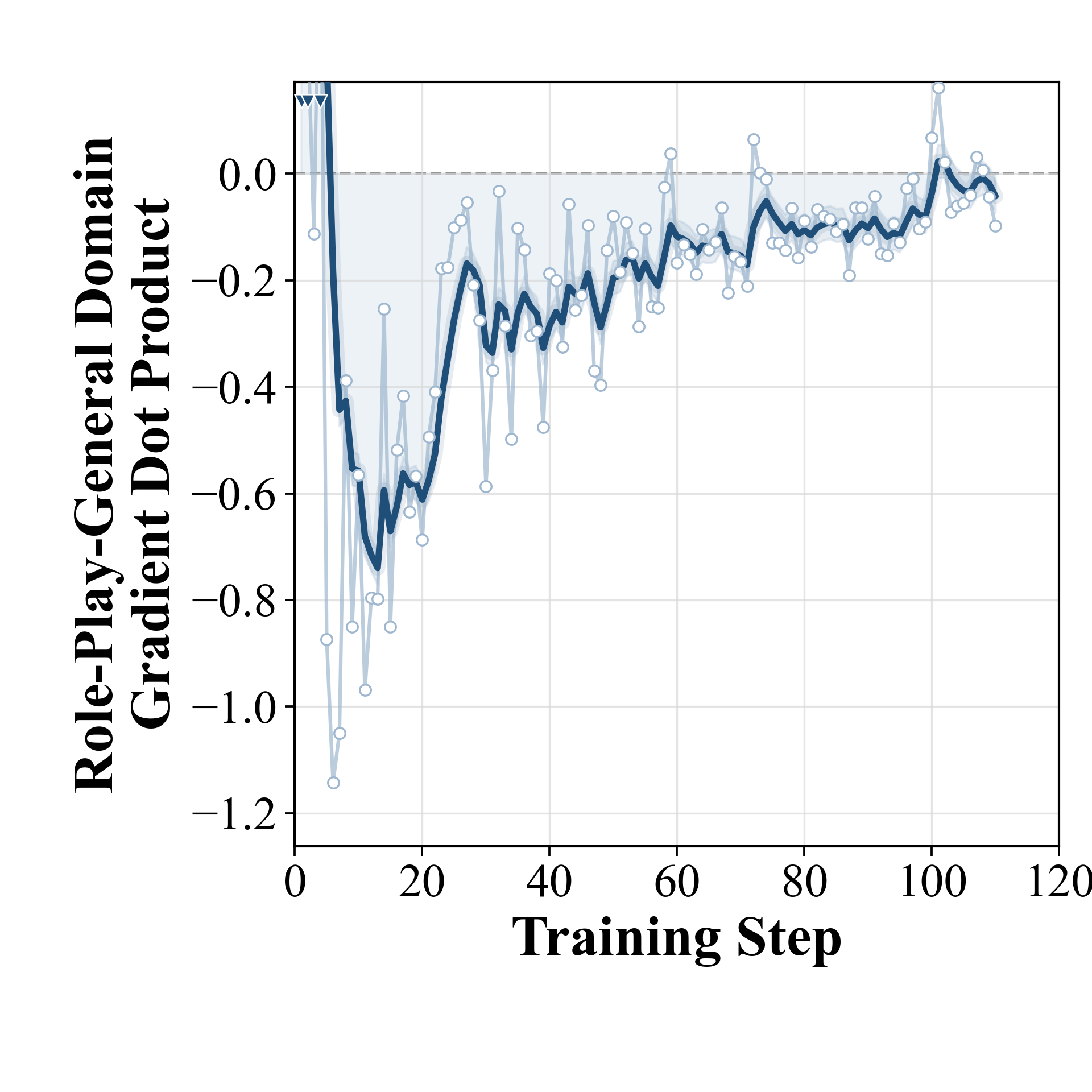}
    \end{subfigure}
    \caption{
        Conceptual overview and training-dynamics.
        \textbf{Top}: Two failure modes of MOPD under incomplete-coverage teacher feedback: recovery-preservation \textbf{counteraction} and weak-signal \textbf{flattening} within a feedback source.
        \textbf{Bottom}: Training dynamics of cross-domain gradient dot products for Medical Reasoning QA and Role-Play Dialogue. \textbf{The predominantly negative values empirically support the recovery-preservation counteraction failure mode.}
    }
    \label{fig:intro_overview}
\end{figure}

To avoid over-reliance on costly collection of the teacher's original data, we consider using proxy prompts to recover general capability. Specifically, starting from an open-source general model, we post-train it into a vertical-domain model, after which MOPD is applied to recover its lost general capability: the original open-source model serves as the general teacher, and the specialized model serves as both the student initialization and the domain teacher. In this situation, domain prompts are available from the specialization pipeline, while the general teacher's original post-training data are unknown. Instead of attempting to reconstruct or cover this hidden distribution, we use readily available public general-domain prompts as \textbf{proxy prompts}, which may incompletely cover the open-source teacher's true training distribution. Under this incomplete-coverage setting, we identify two MOPD failure modes, as illustrated in Figure \ref{fig:intro_overview}: \textbf{recovery-preservation counteraction}, where gradients for general capability recovery and domain behavior preservation point in different directions and counteract when averaged in the same update, as reflected by predominantly negative cross-domain gradient dot products during training; and \textbf{weak-signal flattening}, where a fraction of gradients from different prompts under the same teacher are small and directionally dispersed, flattening the large-norm gradients that are most useful for training.

In this paper, we propose \textbf{Counteraction-Aware Multi-Teacher On-Policy Distillation} (\textbf{CaMOPD}) to address these two failure modes. \textbf{First}, inspired by task-level alternating optimization in multi-task learning~\citep{luong2016multitask, akbari2023alternating}, CaMOPD introduces \textbf{decoupled alternating training} to address recovery-preservation counteraction: by default, it performs three general-recovery steps followed by one domain-preservation step. This alternating schedule allocates dedicated updates to general recovery while periodically preserving domain behavior, thereby mitigating gradient counteraction between recovery and preservation. \textbf{Second}, CaMOPD uses \textbf{gap-based sample selection}. It selects samples with higher rankings in token-level teacher-student log-probability gaps. This concentrates updates on prompts with stronger correction demand, and our gradient coherence analysis shows that the selected subsets produce more aligned update directions. We validate CaMOPD in two stylistically different vertical-domain settings, role-play dialogue and medical reasoning QA. Evaluations show that, under the premise of preserving the acquired domain behavior, CaMOPD outperforms all baselines in general capability recovery. Furthermore, the success across such distinct scenarios verifies the broad applicability and generalization of our method.

Our contributions are summarized as follows:
\begin{itemize}
    \item We identify two MOPD failure modes under proxy prompts: \textbf{recovery-preservation counteraction} and \textbf{weak-signal flattening}.
    \item We propose \textbf{CaMOPD}, integrating \textbf{decoupled alternating training} and \textbf{gap-based sample selection} to address gradient counteraction and concentrate high-demand correction signals.
    \item Evaluations across two \textbf{stylistically distinct} vertical domains validate our analysis, demonstrating the effectiveness of CaMOPD.
\end{itemize}


\section{Related work}

\subsection{On-Policy Distillation}

On-policy distillation (OPD) queries a teacher on student-generated outputs, making it a natural fit for LLM post-training on model-generated trajectories~\citep{gkd, minillm, qwen3_report, deepseek_v4_report}. However, OPD signals are not uniformly useful across trajectories or tokens. One line of work studies signal imbalance, teacher-student compatibility, and noisy or low-coherence teacher feedback, showing that OPD can introduce unreliable update directions~\citep{rethinking_opd, revisiting_opd, skd, paced}. Another line improves OPD by calibrating, relaxing, selecting, or reweighting local supervision at the token or sample level~\citep{scope_opd, reopold, prefix_opd, token_selective_dual_kd, selectkd, adakd, paced}. Unlike these methods that refine local signals, CaMOPD introduces gap-based sample selection to explicitly improve cross-sample gradient coherence.

\subsection{Multi-Teacher On-Policy Distillation}

Multi-teacher OPD (MOPD) extends OPD by using multiple teachers as token-level guidance sources, and has been used in several recent LLM post-training pipelines. MiMo-V2-Flash and DeepSeek-V4 use MOPD-style distillation for base-model post-training and broad capability integration~\citep{mimo_mopd, deepseek_v4_report}. Baichuan-M3, Nemotron-Cascade 2, and GLM-5 apply related ideas to vertical-domain or staged post-training, with the latter two using cross-stage distillation to integrate or recover capabilities that degrade across training stages~\citep{baichuan_m3, nemotron_cascade2, glm5}. Unlike these pipelines that assume fully aligned teacher data, CaMOPD targets an incomplete-coverage setting with proxy prompts, focusing on reducing cross-domain gradient conflicts and elevating signal coherence.

\subsection{Mitigating Catastrophic Forgetting}

Capability degradation during specialization is closely related to catastrophic forgetting in continual adaptation~\citep{continual_finetuning_forgetting}. Existing mitigation strategies commonly mix domain data with general-domain or pretraining data~\citep{sft_data_composition, pretraining_data_injection}, or use replay mechanisms to revisit prior-task or synthesized examples~\citep{gem, experience_replay_cl, replay_processes_cl, self_synthesized_rehearsal}. CaMOPD shares the capability-preservation objective of replay-based approaches, while organizing recovery and preservation through separate alternating updates.


\begin{figure*}[t]
    \centering
    \includegraphics[width=0.95\textwidth]{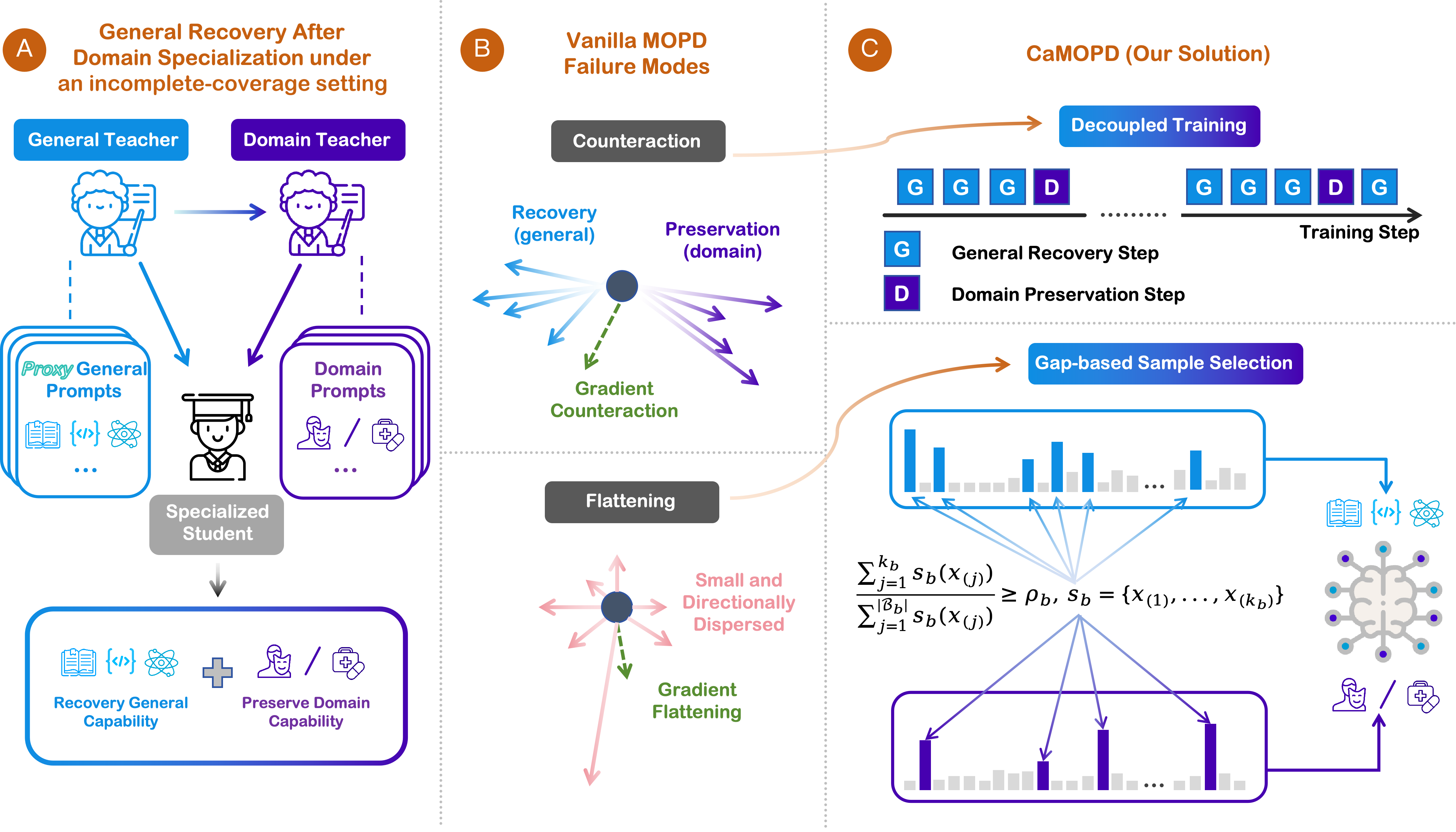}
    \caption{
        Overview of the general capability recovery after domain specialization problem, the Vanilla MOPD failure modes, and the CaMOPD design.
        \textbf{A}: General capability recovery after domain specialization under an \textbf{incomplete-coverage setting}. The original open-source model acts as the general teacher on proxy general prompts, while the specialized model serves as both the student and the domain teacher on domain prompts.
        \textbf{B}: Vanilla MOPD mixes both feedback sources in every update, which can cause recovery-preservation \textbf{counteraction} when general and domain gradients are misaligned, and weak-signal \textbf{flattening} when low-demand or dispersed samples dilute concentrated corrections within a branch.
        \textbf{C}: CaMOPD addresses these issues with decoupled alternating training, together with gap-based sample selection that concentrates each branch on samples with larger teacher-student correction demand.
    }
    \label{fig:core_overview}
\end{figure*}

\section{Method}
\label{sec:method}

This section aims to address the general capability recovery after domain specialization problem under a practical incomplete-coverage setting. As illustrated in Figure~\ref{fig:core_overview}, Part A depicts the proxy-prompt training setup (detailed in Section \ref{subsec:preliminaries}); Part B illustrates the recovery-preservation counteraction and weak-signal flattening failure modes of Vanilla MOPD (analyzed in Section \ref{subsec:failure_modes}); and Part C presents our CaMOPD method, which resolves these issues via decoupled alternating training and gap-based sample selection (introduced in Section \ref{subsec:camopd}).

\subsection{Preliminaries}
\label{subsec:preliminaries}

\paragraph{Problem Setup.} As illustrated in Figure~\ref{fig:core_overview} (Part A), we consider general capability recovery after domain specialization under an incomplete-coverage setting. Given a domain-specialized model, we aim to recover its general capabilities while preserving domain behaviors using proxy general prompts $\mathcal{D}_g$ and domain prompts $\mathcal{D}_d$. The original open-source model serves as the general teacher $T_g$, while the domain-specialized model serves as both the domain teacher $T_d$ and the initialization for the target student policy $\pi_\theta$.

\paragraph{Technical Formulation of MOPD.} Under the multi-teacher on-policy distillation (MOPD) paradigm~\citep{mimo_mopd}, the training student policy $\pi_\theta$ is optimized on rollouts $y=(y_1,\ldots,y_L)$ of length $L$ generated by the inference policy $\mu_\theta$ given prompt $x \sim \mathcal{D}_b$ ($b\in\{g,d\}$). Here, $\mu_\theta$ denotes the inference policy obtained by loading the current training parameters $\theta$ into the inference engine. Let $h_t=(x,y_{<t})$. The token-level teacher feedback defines the advantage function $\hat{A}_{b,t}(\theta)$ as the teacher-student log-probability gap:
\begin{equation}
    \hat{A}_{b,t}(\theta)
    =
    \mathrm{sg}
    \left[
    \log \frac{\pi_{T_b}(y_t\mid h_t)}{\pi_{\theta}(y_t\mid h_t)}
    \right],
    \label{eq:mopd_gap}
\end{equation}
where $\mathrm{sg}[\cdot]$ denotes the stop-gradient operation. The branch-wise Vanilla MOPD loss uses the training--inference importance weight $w_t(\theta)$, defined as $\mathrm{sg}[\pi_\theta(y_t\mid h_t)/\mu_\theta(y_t\mid h_t)]$ when the ratio lies in $[\epsilon_{\mathrm{low}},\epsilon_{\mathrm{high}}]$, and as $0$ otherwise:
\begin{multline}
    \mathcal{L}_{\mathrm{MOPD}}^{b}(\theta)
    =
    -
    \mathbb{E}_{x \sim \mathcal{D}_b, y \sim \mu_\theta(\cdot \mid x)}
    \Bigg[ \\
    \frac{1}{L}
    \sum_{t=1}^{L}
    w_t(\theta) \hat{A}_{b,t}(\theta) \log \pi_\theta(y_t \mid h_t)
    \Bigg],
    \label{eq:mopd_loss}
\end{multline}

\subsection{Failure Modes of Vanilla MOPD}
\label{subsec:failure_modes}

We identify two optimization failure modes that Vanilla MOPD can induce under the incomplete-coverage setting.

\paragraph{Recovery-Preservation Counteraction.} The first failure mode occurs between the two data streams. Vanilla MOPD unconditionally aggregates gradients from the general-prompt and domain-prompt branches during every parameter update. The training dynamics of Vanilla MOPD show that the general-recovery and domain-preservation gradients generally point in conflicting directions, as reflected by predominantly negative cross-domain gradient dot products in Figure~\ref{fig:intro_overview}. Consequently, part of the recovery direction is implicitly counteracted by the preservation direction within the same mixed update.

\paragraph{Weak-Signal Flattening.} The second failure mode occurs within each feedback source. Full-batch averaging assigns equal aggregation weight to all sampled examples, allowing low-demand samples to dilute the contribution of high-demand samples. More critically, examples with weak effective learning signals often exhibit poor cross-sample gradient coherence, so their dispersed directions further weaken the coherent gradients that are most useful for training.

\subsection{Counteraction-Aware MOPD (CaMOPD)}
\label{subsec:camopd}

To resolve the aforementioned bottlenecks, we introduce a scheduled and score-gated distillation strategy that decouples cross-domain aggregation and refines within-branch correction density while fully preserving the MOPD importance sampling core.

\paragraph{Decoupled Alternating Training.} Inspired by task-level alternating optimization in multi-task learning~\citep{luong2016multitask,akbari2023alternating}, CaMOPD addresses recovery-preservation counteraction by assigning the two signals to separate updates under a periodic training schedule:
\begin{equation}
    \mathcal{S}
    =
    \underbrace{G,\ldots,G}_{n_g}, D,
\end{equation}
where $n_g$ general recovery steps ($G$) are followed by a single domain preservation step ($D$). By updating only the active branch $b=\mathcal{S}(t)$ at each step, general and domain gradients are computed at different parameter points, reducing direct mixed-update counteraction between recovery and preservation signals.

\paragraph{Role-specific Gap-based Sample Selection.} To mitigate weak-signal flattening, CaMOPD selects samples based on their correction demand within the active branch. For each rollout $y_i$ of length $L_i$, let $h_{i,t}=(x_i,y_{i,<t})$. We compute the initial token-level gap before the parameter update, $\Delta_{b,i,t} = \log \pi_{T_b}(y_{i,t}\mid h_{i,t}) - \log \pi_\theta(y_{i,t}\mid h_{i,t})$, as a label-free indicator.

General recovery uses the token-average absolute gap to measure deviation from the general teacher:
\begin{equation}
    s_g(x_i)
    =
    \frac{1}{L_i}\sum_{t=1}^{L_i}|\Delta_{g,i,t}|.
    \label{eq:recovery_gap}
\end{equation}
Conversely, domain preservation uses the token-average positive gap to estimate preservation demand:
\begin{equation}
    s_d(x_i)
    =
    \frac{1}{L_i}\sum_{t=1}^{L_i}[\Delta_{d,i,t}]_+.
    \label{eq:preserve_gap}
\end{equation}
Here, $[z]_+=\max(z,0)$ denotes the positive part of $z$.
This asymmetric scoring reflects the different roles of the two branches: general recovery treats any large teacher-student deviation as a useful recovery signal, whereas domain preservation prioritizes tokens that the domain teacher assigns higher likelihood than the current student, avoiding over-penalization of already plausible domain-specific behaviors.
Given a batch $\mathcal{B}_b$ from the active branch $b$ and a gap-score mass target $\rho_b\in(0,1]$, we sort samples by $s_b(x_i)$ in descending order, yielding $x_{(1)},\ldots,x_{(|\mathcal{B}_b|)}$. For $k\in\{1,\ldots,|\mathcal{B}_b|\}$, we choose the smallest prefix length $k_b$ that covers the target cumulative gap-score mass:
\begin{equation}
    k_b
    =
    \min
    \left\{
    k:
    \frac{
    \sum_{j=1}^{k}s_b(x_{(j)})
    }{
    \sum_{j=1}^{|\mathcal{B}_b|}s_b(x_{(j)})
    }
    \geq \rho_b
    \right\}.
    \label{eq:mass_selection}
\end{equation}
The selected subset is then the corresponding prefix $S_b=\{x_{(1)},\ldots,x_{(k_b)}\}$, and the loss calculation is performed on this subset. To intensify the optimization drive on these prioritized instances, we introduce a branch-specific scaling ratio $r_b$ (e.g., $r_g=2, r_d=1$) to scale their generated teacher feedback. For sample $x_i$, let $w_{i,t}(\theta)$ and $\hat{A}_{b,i,t}(\theta)$ denote the corresponding token-level importance weight and advantage, respectively. The final CaMOPD objective for active branch $b$ is formulated as:
\begin{multline}
    \mathcal{L}_{b}^{\mathrm{CaMOPD}}(\theta)
    =
    -
    \frac{1}{|S_b|}
    \sum_{x_i\in S_b}
    \Bigg[ \\
    \frac{1}{L_i}
    \sum_{t=1}^{L_i}
    w_{i,t}(\theta) r_b \hat{A}_{b,i,t}(\theta) \log \pi_{\theta}(y_{i,t}\mid h_{i,t})
    \Bigg].
    \label{eq:camopd_branch_loss}
\end{multline}
The complete end-to-end procedure is summarized in Algorithm~\ref{alg:camopd}.

\begin{algorithm}[htbp]
\small
\linespread{1.08}\selectfont
\caption{Counteraction-Aware Multi-Teacher On-Policy Distillation (CaMOPD)}
\label{alg:camopd}
\begin{algorithmic}[1]
\setlength{\itemsep}{2pt}
\Require Initial student $\pi_{\theta_0}$, teachers $\{\pi_{T_g},\pi_{T_d}\}$, data streams $\{\mathcal{D}_g,\mathcal{D}_d\}$, schedule $\mathcal{S}$, number of training steps $N$, learning rate $\eta$, gap-score mass targets $\{\rho_g,\rho_d\}$, recovery scale $r_g$
\For{training step $t=0,\ldots,N-1$}
    \State Set active branch $b \gets \mathcal{S}(t)$
    \State Set branch scale $r_b\gets r_g$ if $b=g$, otherwise $r_b\gets 1$
    \State Sample prompts $\mathcal{B}_b\sim\mathcal{D}_b$
    \State Load $\theta_t$ into the inference engine to obtain $\mu_{\theta_t}$
    \State Generate on-policy responses $y_i\sim\mu_{\theta_t}(\cdot\mid x_i)$ for $x_i\in\mathcal{B}_b$
    \State Compute initial token gaps $\Delta_{b,i,\tau}=\log\pi_{T_b}(y_{i,\tau}\mid h_{i,\tau})-\log\pi_{\theta_t}(y_{i,\tau}\mid h_{i,\tau})$
    \If{$b=g$}
        \State Score each sample by $s_i\gets \frac{1}{L_i}\sum_{\tau=1}^{L_i}|\Delta_{g,i,\tau}|$
    \Else
        \State Score each sample by $s_i\gets \frac{1}{L_i}\sum_{\tau=1}^{L_i}[\Delta_{d,i,\tau}]_+$
    \EndIf
    \State Sort samples by $s_i$ in descending order
    \State Select the smallest prefix length $k_b$ covering $\rho_b$ of cumulative gap-score mass and set $S_b=\{x_{(1)},\ldots,x_{(k_b)}\}$
    \State Compute $\mathcal{L}_{b}^{\mathrm{CaMOPD}}(\theta_t)$ on $S_b$ using Eq.~\eqref{eq:camopd_branch_loss}
    \State Update $\theta_{t+1}\gets\theta_t-\eta\nabla_\theta\mathcal{L}_{b}^{\mathrm{CaMOPD}}(\theta_t)$
\EndFor
\end{algorithmic}
\end{algorithm}

\begin{table*}[t]
  \caption{Results for the Role-Play dialogue instantiation on general capability benchmarks and Role-Play domain benchmarks. Higher is better for all metrics. Bold marks the best result among baselines. CaMOPD denotes our method.}
  \label{tab:main_results}
  \centering
  \begingroup
  \scriptsize
  \setlength{\tabcolsep}{2pt}
  \renewcommand{\tabularxcolumn}[1]{m{#1}}
  \newcolumntype{Y}{>{\centering\arraybackslash}X}
  \begin{tabularx}{\textwidth}{@{}l*{15}{Y}@{}}
    \toprule
    \textbf{Model}
    & \multicolumn{10}{c}{\textit{General capability}}
    & \multicolumn{5}{c}{\textit{Role-Play domain capability}} \\
    \cmidrule(lr){2-11}\cmidrule(lr){12-16}
    & \shortstack[c]{\textbf{GPQA}\\\textbf{Diamond}}
    & \shortstack[c]{\textbf{Zebra}\\\textbf{Logic}}
    & \shortstack[c]{\textbf{HMMT}\\\textbf{25}}
    & \textbf{LCB v5}
    & \textbf{LCB v6}
    & \shortstack[c]{\textbf{IF}\\\textbf{Eval}}
    & \shortstack[c]{\textbf{Arena}\\\textbf{HP}}
    & \shortstack[c]{\textbf{Arena}\\\textbf{CW}}
    & \shortstack[c]{\textbf{LiveBench}\\\textbf{1125}}
    & \textbf{Avg.}
    & \shortstack[c]{\textbf{Story}\\\textbf{Cons.}}
    & \textbf{Anthro.}
    & \shortstack[c]{\textbf{Char.}\\\textbf{Fid.}}
    & \shortstack[c]{\textbf{Story}\\\textbf{Qual.}}
    & \textbf{Avg.} \\
    \midrule
    \textbf{General Teacher}
    & 61.49 & 79.50 & 33.12 & 34.77 & 20.57 & 83.18 & 36.50 & 65.80 & 65.60 & 53.39 & 34.74 & 23.09 & 23.89 & 41.98 & 30.92 \\
    \textbf{Role-Play Teacher}
    & 54.29 & 31.50 & 26.67 & 25.09 & 16.00 & 67.65 & 9.70 & 4.20 & 42.80 & 30.88 & 42.27 & 33.42 & 38.75 & 49.89 & 41.08 \\
    \midrule
    \textbf{Off-policy KD}
    & 60.35 & 68.50 & 28.54 & 30.47 & 23.43 & 75.97 & 23.60 & 11.20 & 57.00 & 42.12 & 41.25 & 32.15 & 38.23 & 49.61 & 40.31 \\
    \textbf{Vanilla MOPD}
    & 61.99 & 71.30 & 29.17 & 31.54 & 22.29 & 79.67 & 25.50 & 13.70 & 58.90 & 43.78 & 43.57 & \textbf{34.91} & 41.45 & 51.50 & 42.86 \\
    \textbf{Relaxed OPD}
    & \textbf{63.51} & 71.40 & \textbf{31.25} & 32.26 & 22.86 & 79.67 & 30.10 & 17.30 & 59.00 & 45.26 & 45.42 & 33.35 & 41.53 & 52.84 & 43.28 \\
    \textbf{SelecTKD}
    & 60.98 & 70.60 & 31.04 & 34.41 & \textbf{24.00} & 78.93 & 27.10 & 16.90 & 58.90 & 44.76 & 43.87 & 32.28 & 39.46 & 51.54 & 41.79 \\
    \midrule
    \textbf{CaMOPD}
    & 62.63 & \textbf{76.30} & 31.04 & \textbf{37.28} & \textbf{24.00} & \textbf{80.41} & \textbf{33.20} & \textbf{33.30} & \textbf{63.50} & \textbf{49.07} & \textbf{47.81} & 33.96 & \textbf{44.65} & \textbf{53.47} & \textbf{45.00} \\
    \bottomrule
  \end{tabularx}
  \endgroup
\end{table*}

\begin{table*}[t]
  \caption{Results for the Medical Reasoning QA instantiation on general capability benchmarks and Medical domain benchmarks. Higher is better for all metrics. Bold marks the best result among baselines. CaMOPD denotes our method.}
  \label{tab:medical_results}
  \centering
  \begingroup
  \scriptsize
  \setlength{\tabcolsep}{2pt}
  \renewcommand{\tabularxcolumn}[1]{m{#1}}
  \newcolumntype{Y}{>{\centering\arraybackslash}X}
  \begin{tabularx}{\textwidth}{@{}l*{13}{Y}@{}}
    \toprule
    \textbf{Model}
    & \multicolumn{10}{c}{\textit{General capability}}
    & \multicolumn{3}{c}{\textit{Medical domain capability}} \\
    \cmidrule(lr){2-11}\cmidrule(lr){12-14}
    & \shortstack[c]{\textbf{GPQA}\\\textbf{Diamond}}
    & \shortstack[c]{\textbf{Zebra}\\\textbf{Logic}}
    & \shortstack[c]{\textbf{HMMT}\\\textbf{25}}
    & \textbf{LCB v5}
    & \textbf{LCB v6}
    & \shortstack[c]{\textbf{IF}\\\textbf{Eval}}
    & \shortstack[c]{\textbf{Arena}\\\textbf{HP}}
    & \shortstack[c]{\textbf{Arena}\\\textbf{CW}}
    & \shortstack[c]{\textbf{LiveBench}\\\textbf{1125}}
    & \textbf{Avg.}
    & \shortstack[c]{\textbf{MedQA}\\\textbf{USMLE}}
    & \shortstack[c]{\textbf{MedXpertQA}\\\textbf{Text}}
    & \textbf{Avg.} \\
    \midrule
    \textbf{General Teacher}
    & 62.50 & 84.80 & 45.00 & 58.42 & 31.43 & 84.66 & 24.80 & 36.50 & 52.10 & 53.36 & 79.03 & 17.22 & 48.13 \\
    \textbf{Medical Teacher}
    & 62.37 & 71.60 & 28.33 & 48.75 & 27.43 & 48.98 & 10.90 & 36.40 & 34.40 & 41.02 & 86.17 & 22.90 & 54.54 \\
    \midrule
    \textbf{Off-policy KD}
    & 60.10 & 70.30 & 34.38 & 50.54 & 24.57 & 48.61 & 13.20 & 32.60 & 41.70 & 41.78 & 85.23 & 22.41 & 53.82 \\
    \textbf{Vanilla MOPD}
    & 60.10 & 72.60 & 33.96 & 51.25 & 24.00 & 48.06 & 14.80 & 33.00 & 42.80 & 42.29 & \textbf{86.02} & 23.55 & 54.79 \\
    \textbf{Relaxed OPD}
    & 61.49 & \textbf{73.20} & 32.92 & 53.76 & 29.14 & 48.61 & 15.40 & 34.90 & 40.70 & 43.35 & 85.94 & \textbf{24.12} & \textbf{55.03} \\
    \textbf{SelecTKD}
    & 58.46 & 72.10 & 33.54 & 54.48 & \textbf{29.71} & 49.17 & 15.20 & \textbf{35.60} & 41.80 & 43.34 & 85.78 & 23.51 & 54.65 \\
    \midrule
    \textbf{CaMOPD}
    & \textbf{61.99} & 72.60 & \textbf{34.58} & \textbf{56.63} & \textbf{29.71} & \textbf{59.89} & \textbf{16.40} & 31.80 & \textbf{49.00} & \textbf{45.84} & 85.78 & 23.63 & 54.71 \\
    \bottomrule
  \end{tabularx}
  \endgroup
\end{table*}

\section{Experiments}

\subsection{Experimental Setup}

\paragraph{Role-Play Dialogue Instantiation.}
We use Qwen3-4B-Instruct-2507 as the base model and general teacher $T_g$. The role-play domain teacher $T_d$ is obtained by supervised fine-tuning the same base model on CoSER~\citep{coser}. During MOPD, the general-recovery branch uses 10K Nemotron~\citep{llama_nemotron} proxy prompts, while the domain-preservation branch uses 10K held-out long-turn CoSER prompts.

\paragraph{Medical Reasoning QA Instantiation.}
We use Qwen3-8B as the base model and general teacher $T_g$. The domain-specialized initialization and domain teacher $T_d$ are both Intelligent-Internet/II-Medical-8B~\citep{ii_medical_8b}, an open-source medical reasoning model post-trained from Qwen3-8B with medical SFT and DAPO RL~\citep{dapo}. During MOPD, the general-recovery branch uses 10K Nemotron proxy prompts, and the domain-preservation branch uses 10K prompt-only medical examples. More data and training details are provided in Appendix~\ref{sec:dataset_appendix}.

\paragraph{Evaluation.}
For general capability, we evaluate knowledge and reasoning with GPQA-Diamond~\citep{gpqa}, ZebraLogic~\citep{zebralogic}, and HMMT25 from MathArena~\citep{matharena}; coding with LiveCodeBench v5 and LiveCodeBench v6~\citep{livecodebench}; instruction following with IF-Eval~\citep{ifeval}; preference-based generation with the Hard Prompt and Creative Writing subsets of Arena-Hard v2~\citep{arena_hard}; and aggregate live evaluation with LiveBench 1125~\citep{livebench}. For role-play capability, following CoSER~\citep{coser}, we evaluate Storyline Consistency, Anthropomorphism, Character Fidelity, and Storyline Quality. For medical reasoning QA capability, we evaluate MedXpertQA Text and MedQA-USMLE 4-option test accuracy. Appendix~\ref{sec:evaluation_appendix} details the scoring protocol for the general and medical reasoning QA benchmarks.

\subsection{Baselines}

For each domain, we compare Vanilla MOPD, CaMOPD, and two OPD methods reproduced under the Vanilla MOPD training mechanism with the same student initialization, teachers, data streams, batch size, rollout budget, optimizer settings, and training steps. We also include Off-policy KD, which applies supervised fine-tuning to mixed responses generated by the general and domain teachers. Relaxed OPD~\citep{reopold} interprets the teacher-student log-likelihood ratio as a token reward and relaxes strict imitation with reward clipping and entropy-based token sampling. SelecTKD~\citep{selectkd} uses a propose-and-verify token selection rule, applying full distillation loss to accepted tokens while masking or down-weighting rejected tokens.

\begin{figure*}[t]
    \centering
    \begin{subfigure}[t]{0.25\linewidth}
        \centering
        \includegraphics[width=\linewidth]{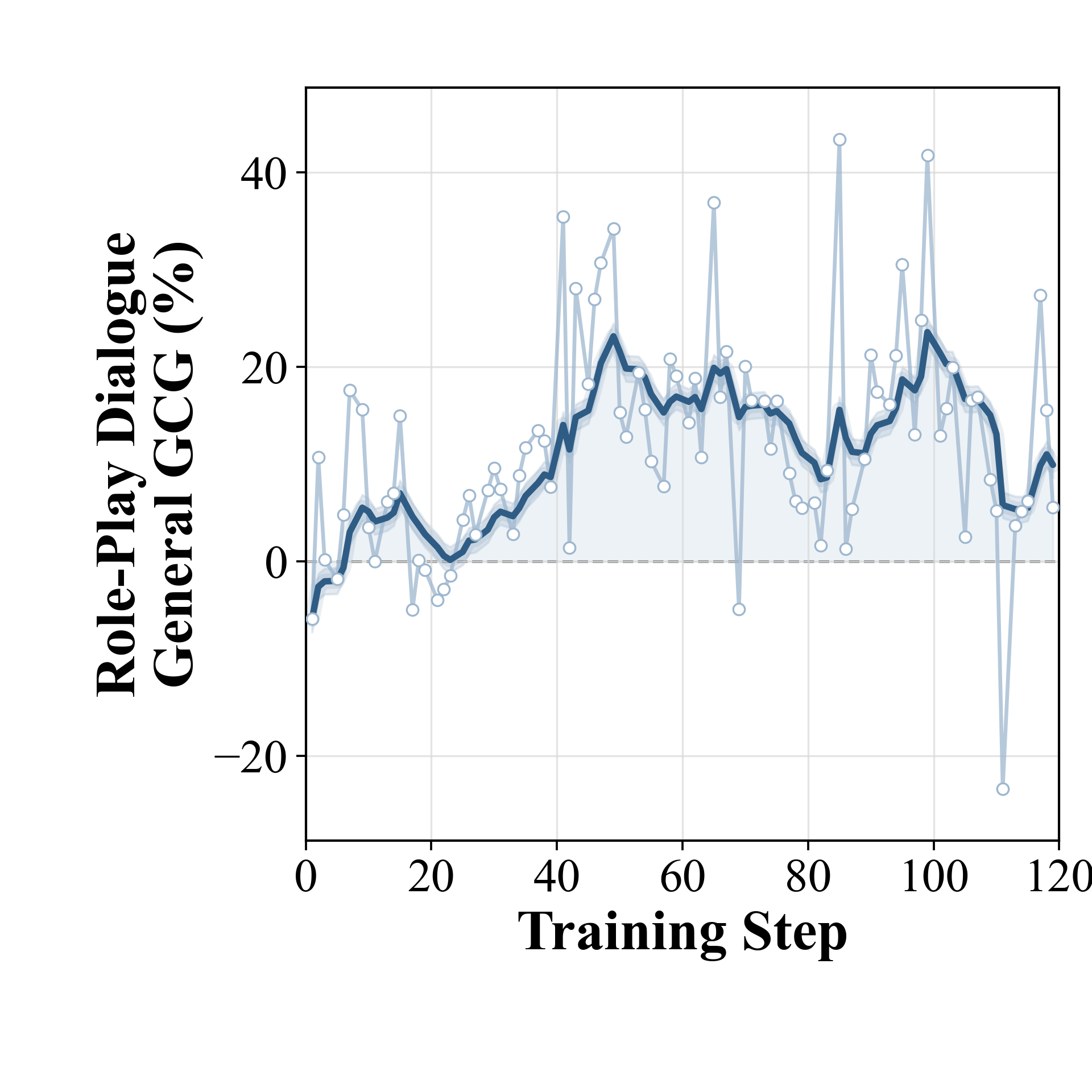}
        \caption{Role-Play Gen. GCG}
        \label{fig:roleplay_gen_gcg}
    \end{subfigure}%
    \begin{subfigure}[t]{0.25\linewidth}
        \centering
        \includegraphics[width=\linewidth]{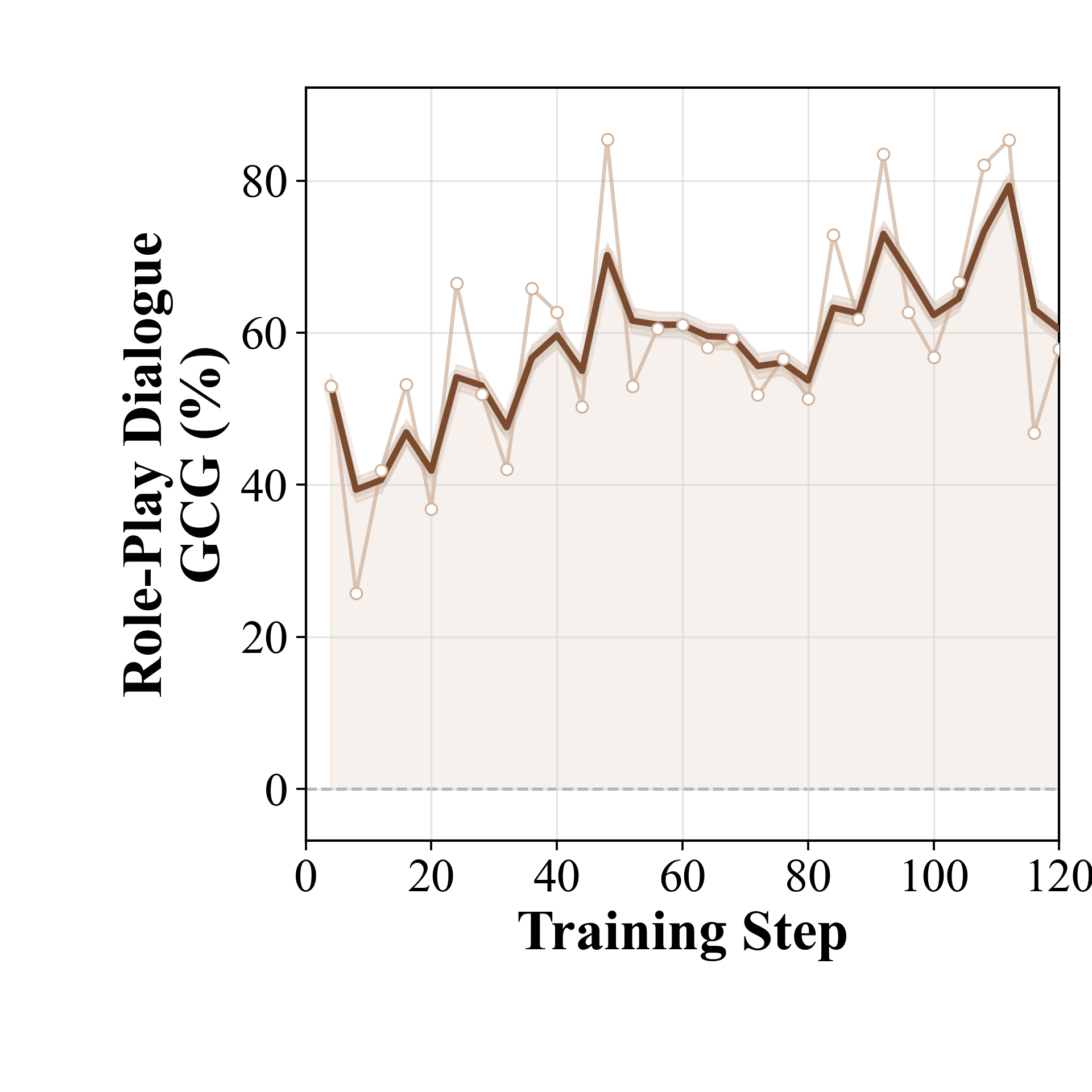}
        \caption{Role-Play Dom. GCG}
        \label{fig:roleplay_dom_gcg}
    \end{subfigure}%
    \begin{subfigure}[t]{0.25\linewidth}
        \centering
        \includegraphics[width=\linewidth]{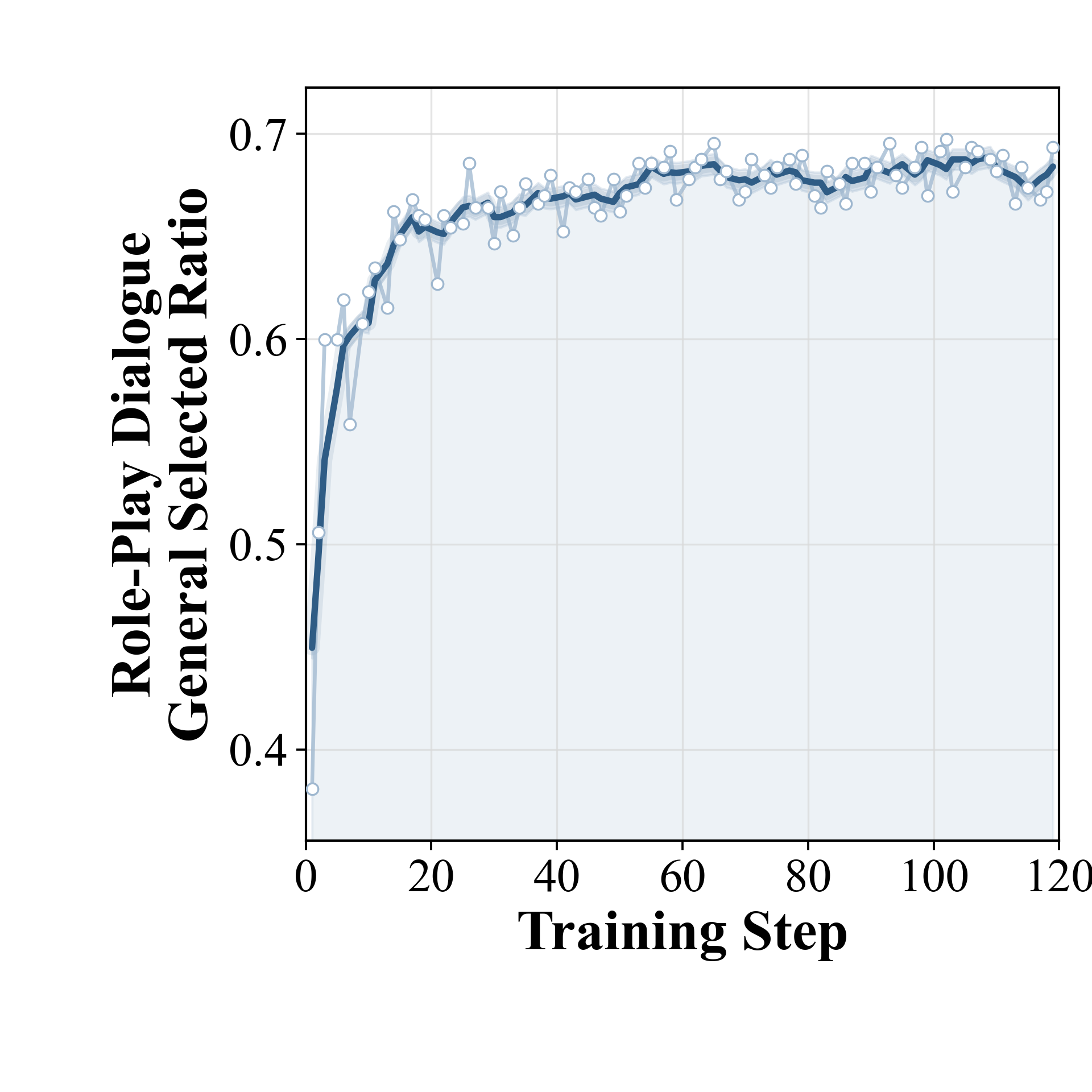}
        \caption{Role-Play Gen. Sel.}
        \label{fig:roleplay_gen_fraction}
    \end{subfigure}%
    \begin{subfigure}[t]{0.25\linewidth}
        \centering
        \includegraphics[width=\linewidth]{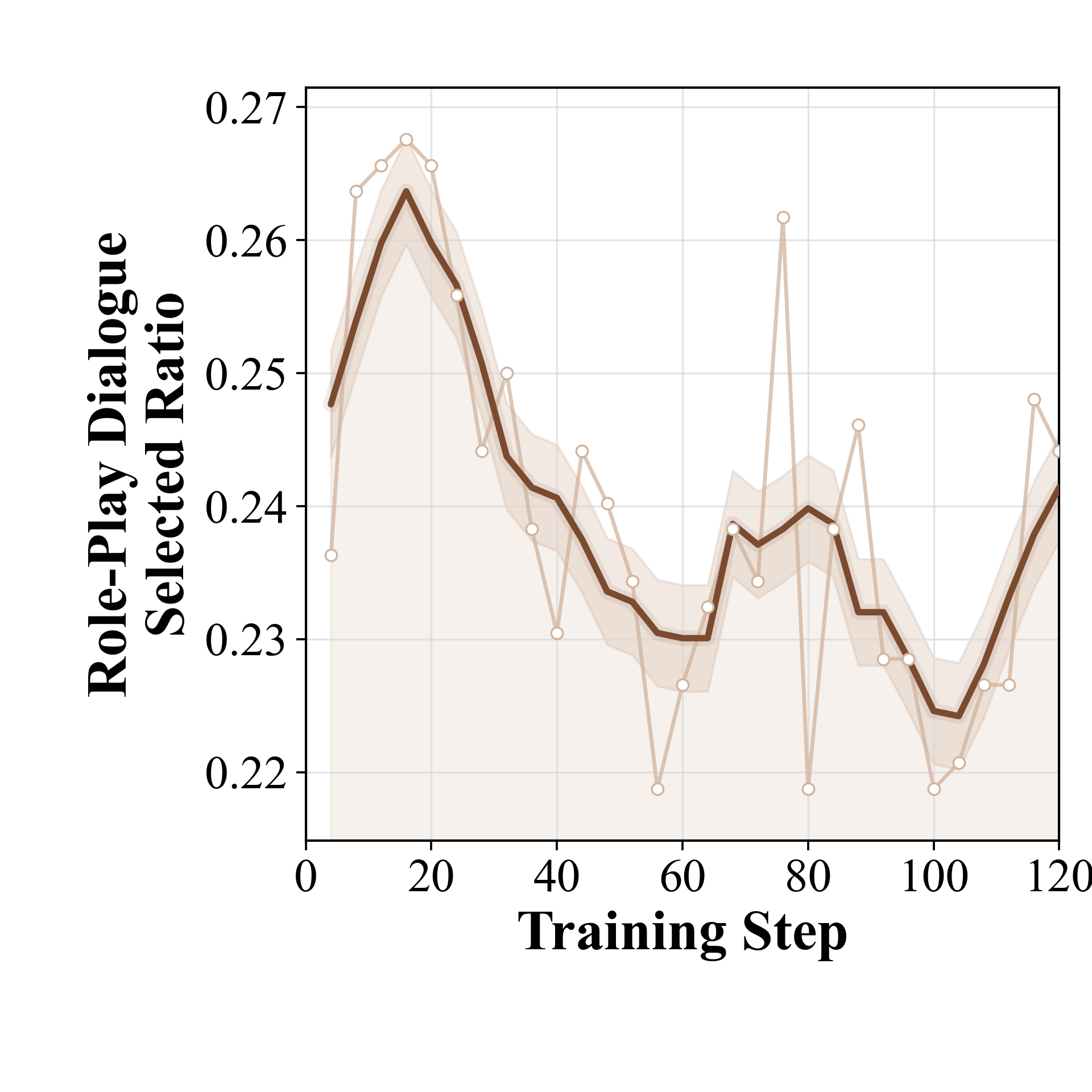}
        \caption{Role-Play Dom. Sel.}
        \label{fig:roleplay_dom_fraction}
    \end{subfigure}

    \begin{subfigure}[t]{0.25\linewidth}
        \centering
        \includegraphics[width=\linewidth]{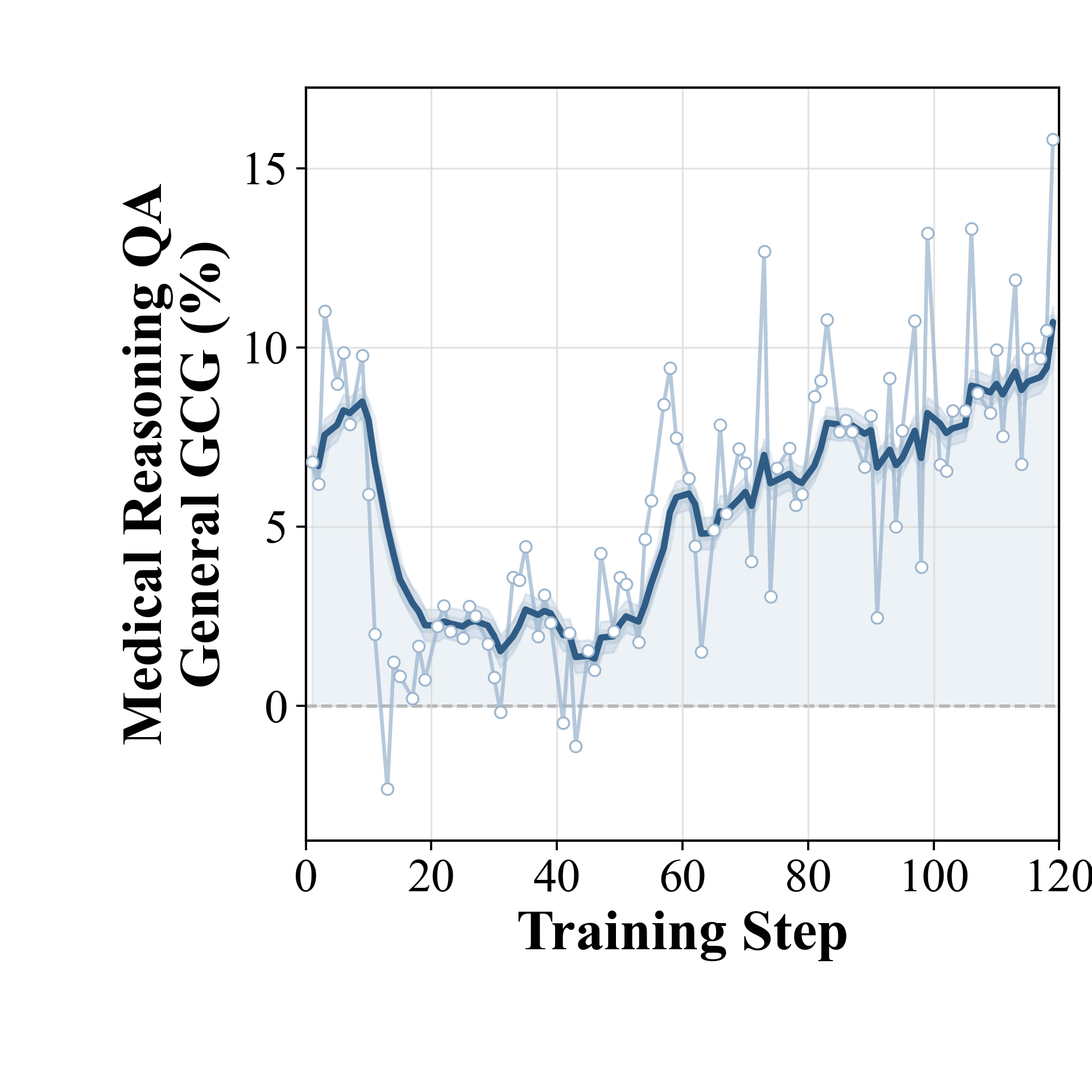}
        \caption{Medical Gen. GCG}
        \label{fig:medical_gen_gcg}
    \end{subfigure}%
    \begin{subfigure}[t]{0.25\linewidth}
        \centering
        \includegraphics[width=\linewidth]{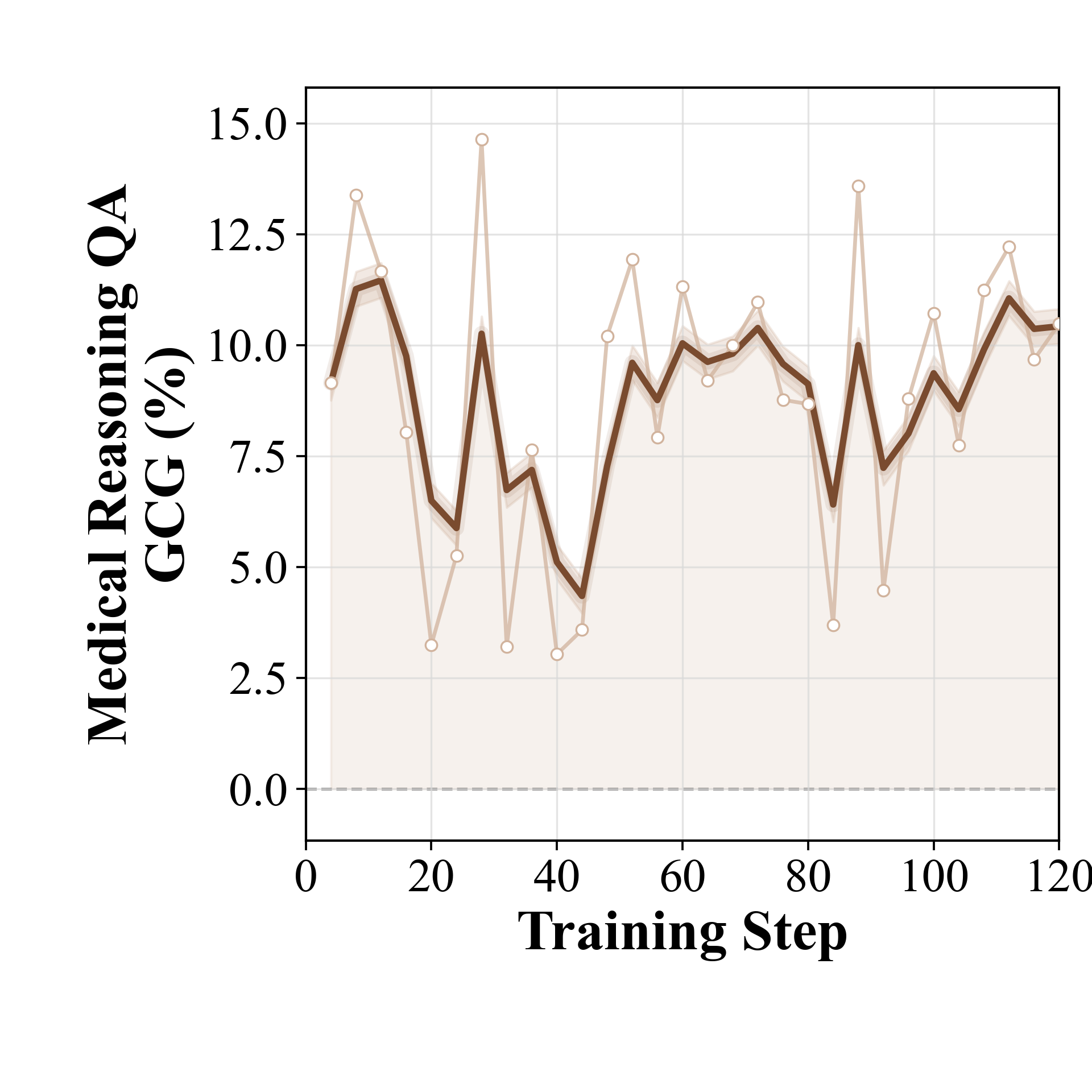}
        \caption{Medical Dom. GCG}
        \label{fig:medical_dom_gcg}
    \end{subfigure}%
    \begin{subfigure}[t]{0.25\linewidth}
        \centering
        \includegraphics[width=\linewidth]{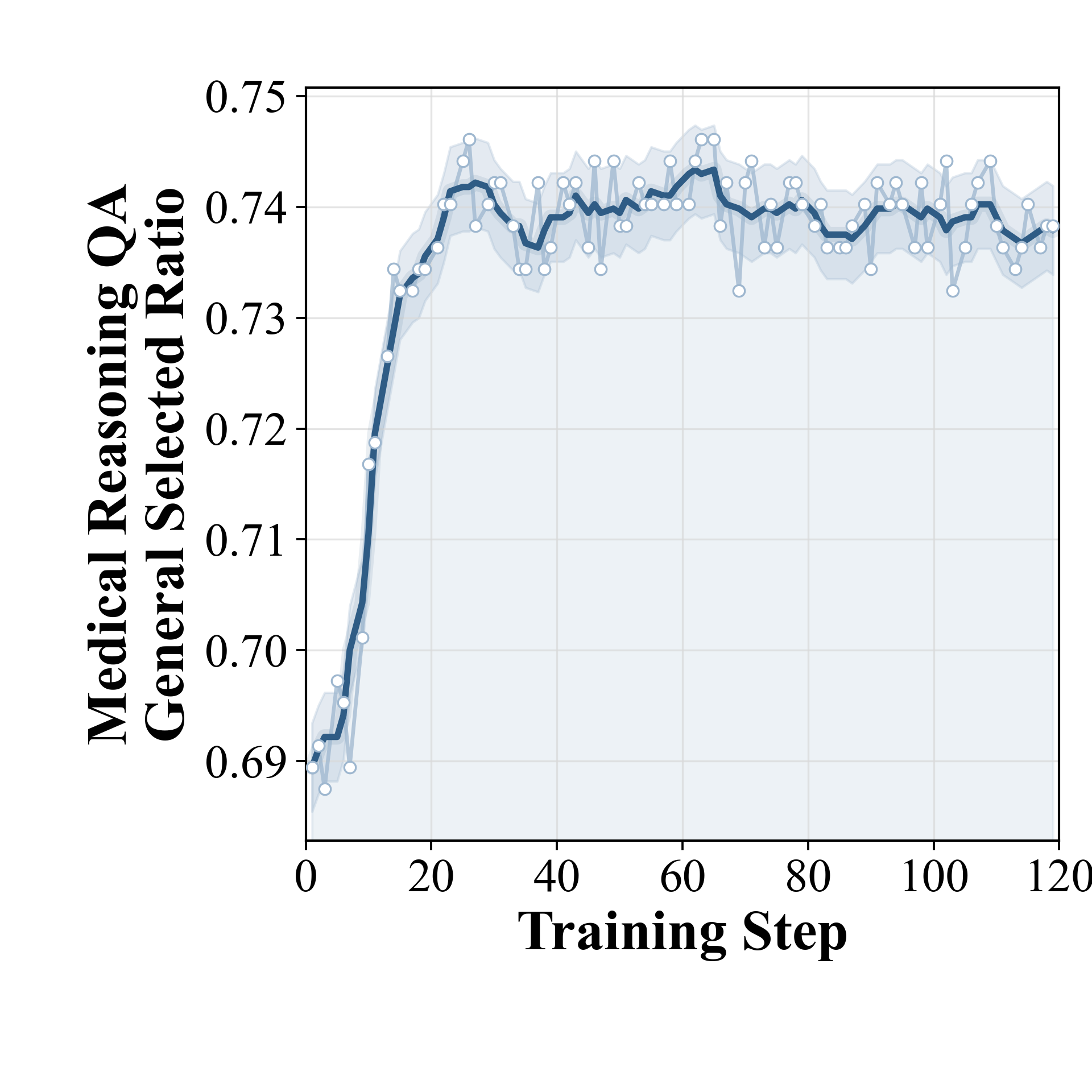}
        \caption{Medical Gen. Sel.}
        \label{fig:medical_gen_fraction}
    \end{subfigure}%
    \begin{subfigure}[t]{0.25\linewidth}
        \centering
        \includegraphics[width=\linewidth]{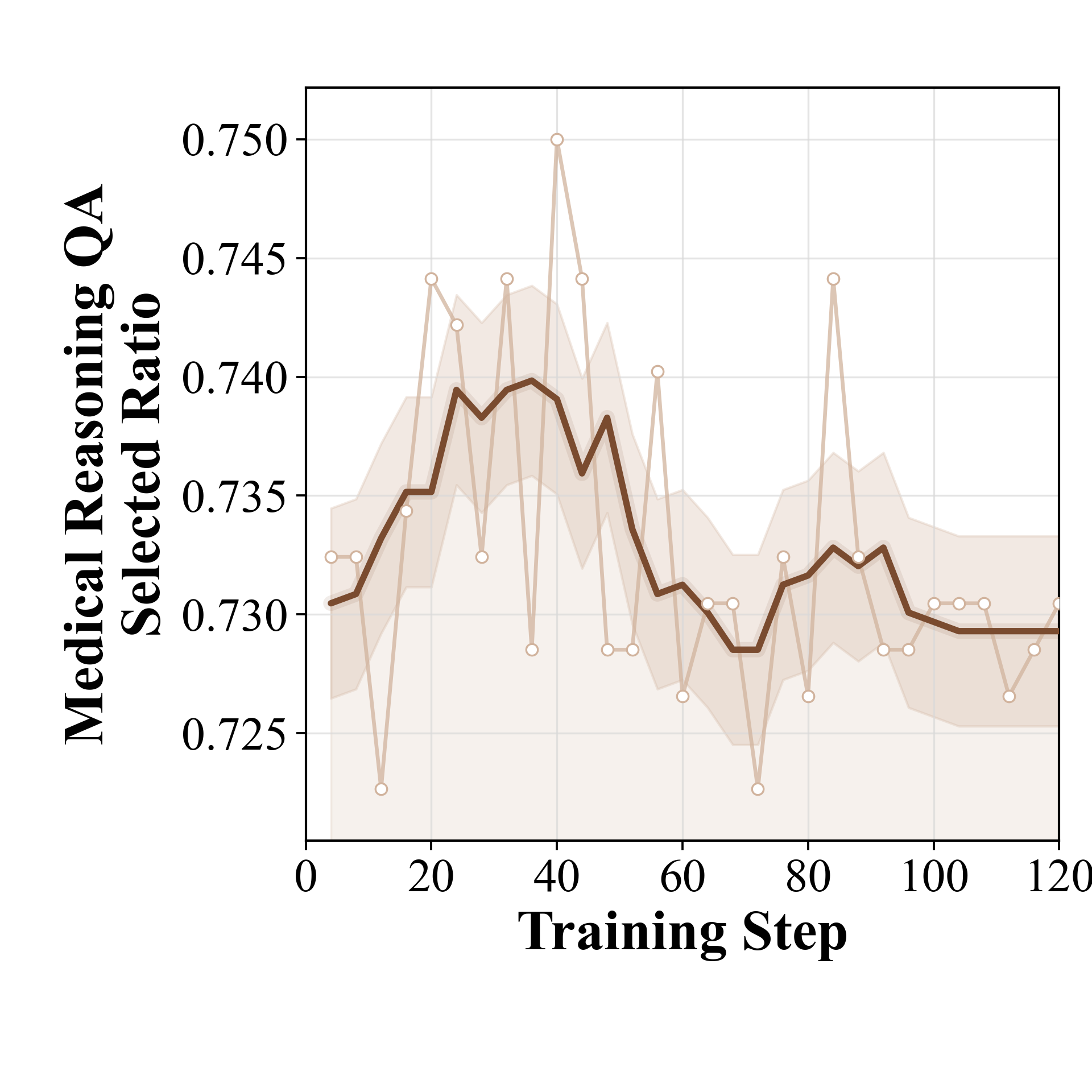}
        \caption{Medical Dom. Sel.}
        \label{fig:medical_dom_fraction}
    \end{subfigure}
    
    \caption{
        Training dynamics and gap-based sample selection analysis. 
        Panels (a)--(d) show Role-Play dialogue, and panels (e)--(h) show Medical Reasoning QA.
        Within each row, the four panels report general-branch GCG, domain-branch GCG, general-branch selected fraction, and domain-branch selected fraction, respectively.
        GCG denotes gradient coherence gain of the selected subset over the full candidate batch, and selected-fraction plots report the retained fraction under the same 80\% gap-score mass target.
    }
    \label{fig:training_dynamics_full}
\end{figure*}

\subsection{Main Results}

Tables~\ref{tab:main_results} and~\ref{tab:medical_results} report the main comparisons for role-play and medical reasoning QA specialization on both general benchmarks and domain-specific evaluations. Across these two stylistically different instantiations, CaMOPD achieves strong general-capability recovery while maintaining the specialized domain behavior, demonstrating the effectiveness and generality of our counteraction-aware design.

\paragraph{General Capability Recovery.}
CaMOPD consistently improves recovery over baselines on the general benchmarks. In the Role-Play dialogue instantiation, CaMOPD raises the general average from the strongest baseline score of 45.26 to 49.07. At the individual-benchmark level, CaMOPD achieves the best or tied-best results on seven of nine general benchmarks. For example, relative to the strongest baseline on each benchmark, it improves Arena-Hard v2 Creative Writing from 17.30 to 33.30, ZebraLogic from 71.40 to 76.30, and LiveBench 1125 from 59.00 to 63.50. In the Medical Reasoning QA instantiation, CaMOPD raises the general average from the strongest baseline score of 43.35 to 45.84. At the individual-benchmark level, CaMOPD achieves the best or tied-best results on seven of nine general benchmarks. For example, relative to the strongest baseline on each benchmark, it improves IF-Eval from 49.17 to 59.89, LiveBench 1125 from 42.80 to 49.00, and LiveCodeBench v5 from 54.48 to 56.63. These results demonstrate the effectiveness of CaMOPD across both specialization settings.

\paragraph{Domain Capability Preservation.}
MOPD-based training preserves the specialized domain behavior across both instantiations. Our method and the MOPD-based baselines all remain close to or above the corresponding domain teachers on the domain metrics, indicating that the MOPD training mechanism itself is sufficient to protect the acquired domain capability.

\subsection{Training Dynamics and Selection Analysis}

We analyze retained sample fractions and within-branch gradient coherence to characterize how gap-based sample selection concentrates updates on samples with larger correction demand and reduces within-branch signal dilution.

\paragraph{Selection Ratio Under The Same Mass Target.}
Figures~\ref{fig:roleplay_gen_fraction}, \ref{fig:roleplay_dom_fraction}, \ref{fig:medical_gen_fraction}, and~\ref{fig:medical_dom_fraction} report the retained sample fractions under the same 80\% gap-score mass target. The general branch requires broad coverage in both domains. The domain branch is sparse in role-play dialogue but remains broader in medical reasoning QA, suggesting that domain correction demand can vary substantially across specialized tasks.

\paragraph{Within-branch Gradient Coherence.}
To examine whether this selection is associated with reduced weak-signal flattening, we measure the within-branch gradient coherence of the parameter updates. For a sample set $A$, we define
\[
    \mathrm{Coh}(A)
    =
    \frac{\left\lVert \sum_{i \in A} g_i \right\rVert}
         {\sum_{i \in A} \left\lVert g_i \right\rVert},
\]
where $g_i$ is the parameter gradient of sample $i$. Higher values indicate more aligned gradient directions. We report gradient coherence gain (GCG) as the relative improvement of the selected subset over the full candidate batch:
\[
\mathrm{GCG}
    =
    \frac{\mathrm{Coh}(A_{\mathrm{selected}}) - \mathrm{Coh}(A_{\mathrm{full}})}
         {\mathrm{Coh}(A_{\mathrm{full}})}
    \times 100\%.
\]

Figures~\ref{fig:roleplay_gen_gcg}, \ref{fig:roleplay_dom_gcg}, \ref{fig:medical_gen_gcg}, and~\ref{fig:medical_dom_gcg} plot the GCG of the selected subset over the full candidate batch. GCG is predominantly positive across both specialization settings, with especially large gains in the role-play domain branch and smaller but consistent gains in medical reasoning QA. Together with the selection-ratio analysis, this suggests that gap-based sample selection reduces weak-signal flattening and produces more coherent update directions across both specialization settings.

\subsection{Hyperparameter Analysis}

\begin{figure}[t]
    \centering
    \includegraphics[width=\linewidth]{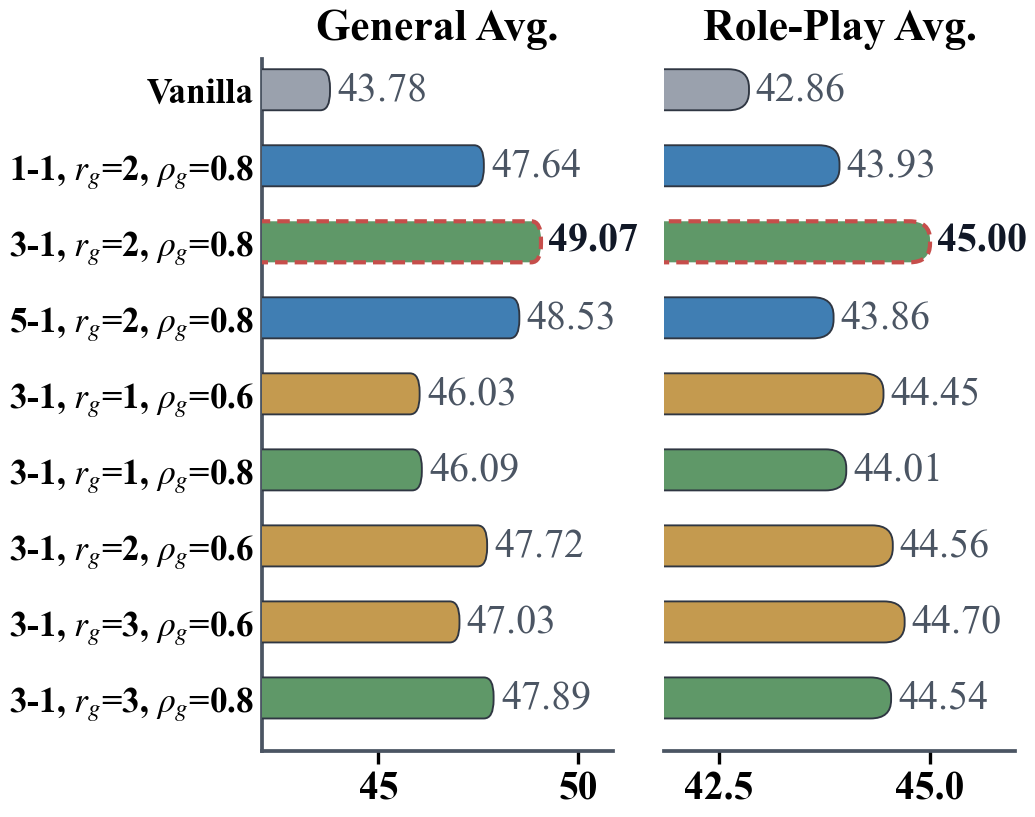}
    \caption{Bar-chart summary of the hyperparameter study. General Avg. is the arithmetic mean over the nine general benchmarks in Table~\ref{tab:hyperparams}; Role-Play Avg. uses the domain-average scores.}
    \label{fig:hyperparameter_results_bar}
\end{figure}

We further examine the recovery-to-preservation schedule, the recovery scaling ratio $r_g$, and the gap-score mass target $\rho_g$ in the role-play setting. As summarized in Figure~\ref{fig:hyperparameter_results_bar}, the 3-1 schedule gives the best recovery-preservation trade-off; the 5-1 schedule achieves similar general recovery, but domain performance starts to decline. For recovery scaling and gap-based sample selection, $r_g=2$ and $\rho_g=0.8$ gives the strongest aggregate general recovery, combining broader sample coverage with amplified gradients on selected high-demand samples. We therefore use the 3-1 schedule with $r_g=2$ and $\rho_g=0.8$ in the main experiments. The full metric table and detailed analysis are provided in Appendix~\ref{sec:hyperparameter_appendix}.

\subsection{Ablation Study}

Table~\ref{tab:ablation_delta} presents a matched ablation chain that successively isolates gap-based sample selection, recovery scaling, and decoupled alternating training across both specialization settings. Each adjacent pair changes one component while keeping the remaining training settings fixed. \textit{CaMOPD} drops the low-gap tail under a 3-1 alternating schedule and applies $r_g=2$ to the high-gap samples; \textit{w/o selection} follows the same schedule, applies $r_g=2$ to the high-gap samples, and leaves the low-gap samples unscaled; \textit{w/o selection and scaling} uses all samples without recovery scaling; and \textit{w/o selection, scaling, and decoupling} uses mixed updates with a 3:1 ratio of general to domain samples. Across both settings, the general and domain averages decrease at each successive ablation. The full per-metric results are provided in Appendix~\ref{sec:ablation_appendix}.

\begin{table}[t]
  \caption{Matched ablations across Role-Play dialogue and Medical Reasoning QA. Gen. Avg. is the arithmetic mean over the nine general benchmarks, and Dom. Avg. is the corresponding domain average. Higher is better, and bold marks the best result.}
  \label{tab:ablation_delta}
  \centering
  \begingroup
  
  \scriptsize
  \setlength{\tabcolsep}{1pt}
  \renewcommand{\arraystretch}{1.08}
  \begin{tabularx}{\columnwidth}{@{}>{\raggedright\arraybackslash}p{0.525\columnwidth}*{4}{>{\centering\arraybackslash}X}@{}}
    \toprule
    \multirow[t]{2}{*}{\textbf{Variant}}
    & \multicolumn{2}{c}{\shortstack[c]{\textbf{Role-Play}\\\textbf{Dialogue}}}
    & \multicolumn{2}{c}{\shortstack[c]{\textbf{Medical}\\\textbf{Reasoning QA}}} \\
    \cmidrule(lr){2-3}\cmidrule(l){4-5}
    & \mbox{\scalebox{0.75}{\textbf{Gen. Avg.}}}
    & \mbox{\scalebox{0.75}{\textbf{Dom. Avg.}}}
    & \mbox{\scalebox{0.75}{\textbf{Gen. Avg.}}}
    & \mbox{\scalebox{0.75}{\textbf{Dom. Avg.}}} \\
    \midrule
    \textbf{CaMOPD} & \textbf{49.07} & \textbf{45.00} & \textbf{45.84} & \textbf{54.71} \\
    \textbf{w/o selection} & 47.21 & 43.46 & 44.94 & 54.61 \\
    \textbf{w/o selection and scaling} & 46.51 & 42.60 & 44.63 & 54.59 \\
    \textbf{w/o selection, scaling, and decoupling} & 46.10 & 42.00 & 43.19 & 54.55 \\
    \bottomrule
  \end{tabularx}
  \endgroup
\end{table}


\section{Conclusion}

We studied general capability recovery after domain specialization under an incomplete-coverage setting, where only proxy general prompts are available for the general teacher. We identified two failure modes of Vanilla MOPD: recovery-preservation counteraction and weak-signal flattening. To address them, we proposed CaMOPD, which combines alternating recovery/preservation updates with gap-based sample selection. Experiments on role-play dialogue and medical reasoning QA show that CaMOPD improves general recovery over baselines while maintaining domain capability. Gradient and selection analyses further suggest that CaMOPD reduces mixed-update counteraction and produces more coherent correction signals.

\section*{Limitations}
Our experiments instantiate CaMOPD with 4B and 8B models in the role-play and medical reasoning QA domains, respectively. Due to computational constraints, we do not evaluate larger model scales in this work; extending CaMOPD to larger specialized models is an important direction for future exploration. In addition, we use a fixed 32K response-length budget for on-policy rollouts. This setting is chosen to avoid artificially limiting the output capacity of long-reasoning models during distillation, but it does not study how different rollout length budgets affect the recovery-preservation trade-off. Future work can examine CaMOPD under shorter and adaptive rollout budgets.

\section*{Ethical Considerations}
This work studies post-training methods for recovering general capabilities while preserving domain-specific behavior. In the medical reasoning QA setting, the models and evaluations are used only for research on medical question answering and educational reasoning benchmarks. The resulting systems should not be interpreted as providing professional medical advice, diagnosis, or treatment recommendations. Any deployment in clinical or patient-facing settings would require substantially stronger validation, expert oversight, safety guardrails, and compliance with applicable medical regulations. Our training and evaluation use publicly available datasets and open-source models. We do not use private patient records or personally identifiable medical information.

\section*{Acknowledgments}
We thank the reviewers for their constructive feedback. We also thank our colleagues on Kuaishou's infrastructure team for their support with the training and evaluation infrastructure.

\bibliography{references}

\clearpage
\appendix

\section{Dataset Appendix: Dataset Details and Processing}
\label{sec:dataset_appendix}

\subsection{Training Sources}

For the role-play experiments, we start from Qwen3-4B-Instruct-2507~\citep{qwen3_report}. We use two teachers during MOPD. The general teacher $T_g$ is the original Qwen3-4B-Instruct-2507 model, and the role-play teacher $T_d$ is obtained by supervised fine-tuning the same base model on CoSER. CoSER~\citep{coser}\footnote{\nolinkurl{https://huggingface.co/datasets/Neph0s/CoSER}} is a role-play dataset built from 771 novels and provides multi-turn, multi-character conversations with character profiles, dialogue context, plot summaries, and character experiences. The official dataset card reports a formatted SFT file, \texttt{sft\_conversations\_sharegpt.json}, together with a held-out test file and full book-level JSON files.

For the medical reasoning QA experiments, we instantiate the same recovery-under-preservation setup with a publicly released medical post-trained model. The general teacher is Qwen/Qwen3-8B~\citep{qwen3_report}, and the domain-specialized initialization and domain teacher are both Intelligent-Internet/II-Medical-8B~\citep{ii_medical_8b}\footnote{\nolinkurl{https://huggingface.co/Intelligent-Internet/II-Medical-8B}}. The model card lists II-Medical-8B under the Apache-2.0 license and describes it as a medical reasoning model obtained by applying medical supervised fine-tuning and DAPO RL optimization~\citep{dapo} on top of Qwen/Qwen3-8B. Accordingly, II-Medical-8B provides the domain-specialized initialization and teacher in this instantiation.

For general-domain recovery, we use Nemotron-Post-Training-Dataset-v1~\citep{llama_nemotron}\footnote{\nolinkurl{https://huggingface.co/datasets/nvidia/Nemotron-Post-Training-Dataset-v1}}, a large SFT corpus released with five splits: chat, code, math, stem, and tool calling. The official card reports 25,659,642 examples in total, with 746,622 chat examples, 1,896,395 code examples, 2,044,407 math examples, 20,662,167 stem examples, and 310,051 tool-calling examples. We sample 10K general prompts proportionally from these categories.

\subsection{CoSER Split for Specialization and Domain Preservation}

We first rank CoSER conversations by the number of dialogue turns. The 10K conversations with the largest number of turns are reserved as unseen role-play domain data for MOPD. For each reserved conversation, we use all but the final turn as the prompt/context and leave the final continuation to be generated on-policy by the current student. These examples are never used in role-play SFT. The remaining 300K CoSER conversations are used to fine-tune Qwen3-4B-Instruct-2507 and produce the role-play teacher.

Using held-out prompts ensures that domain-preservation updates are computed on contexts excluded from role-play SFT. Long-turn conversations require the model to maintain persona, dialogue state, and storyline consistency over rich contexts, providing a challenging preservation setting.

\subsection{Nemotron Sampling for General Recovery}

Because the pretraining and post-training data of the open-source general teacher are unknown, we use Nemotron as a proxy for the general distribution. We sample 10K prompts from its 25.66M examples while preserving the official category proportions, yielding the incomplete-coverage recovery setting studied in our experiments.

\subsection{Medical Reasoning QA Model and Domain-Preservation Prompts}

The medical reasoning QA instantiation uses II-Medical-8B as both the starting student and the preservation teacher. This mirrors the role-play setup at the level of the recovery objective: CaMOPD starts from a domain-specialized model, recovers general capability from the corresponding Qwen3-8B teacher, and periodically performs domain-preservation updates on medical reasoning QA prompts to reduce domain drift.

The medical reasoning QA domain-preservation stream is prompt-only. It does not use gold answers during MOPD/CaMOPD training: the current student first generates on-policy continuations, and the domain teacher provides token-level log-probability feedback on those student trajectories. We construct the preservation prompts from medical question-answering and reasoning sources related to the II-Medical data ecosystem: 4K prompts from MedMCQA~\citep{medmcqa}\footnote{\nolinkurl{https://huggingface.co/datasets/openlifescienceai/medmcqa}}, 3K from MedReason~\citep{medreason}\footnote{\nolinkurl{https://huggingface.co/datasets/UCSC-VLAA/MedReason}}, 2K English prompts from Medical-R1-Distill-Data~\citep{huatuogpt_o1}\footnote{\nolinkurl{https://huggingface.co/datasets/FreedomIntelligence/Medical-R1-Distill-Data}}, and 1K from m23k-tokenized~\citep{m1_medical_reasoning}\footnote{\nolinkurl{https://huggingface.co/datasets/UCSC-VLAA/m23k-tokenized}}. After selection and formatting, the resulting preservation set contains 10K prompts.

\subsection{Data Examples}

Figures~\ref{fig:nemotron_prompt_example}--\ref{fig:medical_prompt_template} show the field structure and training usage for the general, role-play, and medical reasoning QA streams.

\newcommand{\metakey}[1]{\textcolor{magenta!70!black}{\texttt{#1}}}
\newcommand{\rolekey}[1]{\textcolor{cyan!60!black}{\texttt{#1}}}
\newcommand{\datakey}[1]{\textcolor{orange!80!black}{\texttt{#1}}}

\begin{figure}[t]
\centering
\setlength{\fboxsep}{6pt}
\fcolorbox{teal!55!black}{teal!5}{%
\begin{minipage}{0.94\linewidth}
\small
\textbf{Nemotron general prompt example.}
\vspace{2pt}

\metakey{category}: \datakey{math/code/stem/chat/tool\_calling}

\metakey{messages}: \\
\hspace*{0.8em}\rolekey{[user]} a task prompt, such as a \datakey{math problem}, \datakey{programming challenge}, \datakey{STEM question}, \datakey{chat request}, or \datakey{tool-use instruction}.\\
\hspace*{0.8em}\rolekey{[assistant]} a synthetic reference response generated by public/open models.

\metakey{metadata}: source information and, for tool-calling data, tool schema or call metadata.
\end{minipage}}
\caption{Nemotron general prompt structure used for the general recovery branch.}
\label{fig:nemotron_prompt_example}
\end{figure}

\begin{figure}[t]
\centering
\setlength{\fboxsep}{6pt}
\fcolorbox{violet!70!black}{violet!6}{%
\begin{minipage}{0.94\linewidth}
\small
\textbf{CoSER Role-Play conversation example.}
\vspace{2pt}

\metakey{conversations}: \\
\hspace*{0.8em}\rolekey{[system]} role instruction and character/background profile, e.g., ``Your role is to be [character]'' plus \datakey{scenario context}.\\
\hspace*{0.8em}\rolekey{[user/character]} previous dialogue turns and actions.\\
\hspace*{0.8em}\rolekey{[assistant/character]} the next in-character response.

\metakey{auxiliary fields}: \datakey{character profiles}, \datakey{scenario}, \datakey{topic}, \datakey{plot summary}, \datakey{speaking characters}, and \datakey{book-level metadata} in the full CoSER files.
\end{minipage}}
\caption{CoSER Role-Play conversation structure used for Role-Play specialization and domain preservation.}
\label{fig:coser_prompt_example}
\end{figure}

\begin{figure}[t]
\centering
\setlength{\fboxsep}{6pt}
\fcolorbox{orange!70!black}{orange!6}{%
\begin{minipage}{0.94\linewidth}
\small
\textbf{Medical domain-preservation prompt template.}
\vspace{2pt}

\metakey{raw fields}: \datakey{question}, optional \datakey{options} or \datakey{opa/opb/opc/opd}

\metakey{prompt}: \\
\hspace*{0.8em}\rolekey{[system]} You are a careful medical reasoning assistant. Provide educational medical information, mention uncertainty when relevant, and recommend professional medical care for diagnosis or treatment decisions.\\
\hspace*{0.8em}\rolekey{[user]} \datakey{\{question\}}\\
\hspace*{4.8em}\datakey{A. \{opa\}}\\
\hspace*{4.8em}\datakey{B. \{opb\}}\\
\hspace*{4.8em}\datakey{C. \{opc\}}\\
\hspace*{4.8em}\datakey{D. \{opd\}}\\
\hspace*{4.8em}Please reason step by step and give the final answer choice at the end.

\metakey{open-ended variant}: for sources with only \datakey{question} or \datakey{prompt/text}, the user message is \datakey{\{question\}} followed by ``Please reason step by step and answer clearly. Do not present the answer as a substitute for professional medical advice.''

\metakey{data\_type}: \datakey{medical}

\metakey{training use}: gold answers and explanations are not placed in the prompt field; teacher feedback is computed on student-generated continuations.
\end{minipage}}
\caption{Medical domain-preservation prompt template constructed from source dataset fields.}
\label{fig:medical_prompt_template}
\end{figure}

\subsection{Resulting Data Roles}

Table~\ref{tab:dataset_roles} summarizes how each data source is used in the training pipeline. We separate specialization data, domain-preservation prompts, and proxy general-recovery prompts to avoid mixing their roles in the MOPD/CaMOPD setup.

\begin{table*}[t]
  \centering
  \small
  \caption{Dataset roles in CaMOPD. Domain-preservation prompts provide preservation gradients for the active vertical domain, while the general branch uses a compact proportional sample from Nemotron to recover general capability under a limited MOPD budget.}
  \begin{tabular}{llll}
    \toprule
    Data & Size & Used for & Notes \\
    \midrule
    CoSER remainder & 300K & Role-Play SFT teacher & Trains $T_d$ \\
    CoSER long-turn heldout & 10K & Role-Play domain preservation & Unseen by Role-Play SFT \\
    Medical prompt mixture & 10K & Medical domain preservation & Prompt-only; no labels used \\
    Nemotron proportional sample & 10K & General recovery & Proxy general prompts \\
    \bottomrule
  \end{tabular}
  \label{tab:dataset_roles}
\end{table*}

\subsection{Training Protocol}

We perform SFT using the ms-swift framework~\citep{zhao2024swift} and implement the MOPD training pipeline using verl~\citep{sheng2024hybridflow}.

All MOPD-based methods generate one response per prompt with the current student and compute token-level teacher feedback on student trajectories. We train for 120 steps with a prompt batch size of $512$ and a learning rate of $2\times 10^{-6}$. Prompts are truncated at $4096$ tokens and rollouts are capped at $32768$ tokens. Unless otherwise specified, CaMOPD uses a 3-1 recovery-to-preservation schedule, gap-score mass targets of $\rho_g=\rho_d=0.8$, and a general recovery scale of $r_g=2$, with the domain scale fixed at $r_d=1$. We run all training on a $4$-node cluster with $8$ NVIDIA H100 GPUs per node. Each run uses $32$ H100 GPUs for approximately $7$ hours, corresponding to about $224$ total GPU hours.

\paragraph{Gradient Computation.} For the analyses reported in Figures~\ref{fig:intro_overview} and~\ref{fig:training_dynamics_full}, we compute gradients with respect to all model parameters at each analyzed training step and aggregate the resulting statistics across parameter shards under tensor and pipeline parallelism.


\begin{table*}[t]
  \centering
  \small
  \renewcommand{\arraystretch}{1.12}
  \caption{Evaluation settings and scoring protocols for the general benchmarks. \texttt{T} denotes the evaluated model's generation temperature, $n$ denotes the number of generated samples, and $k$ denotes the pass@$k$ aggregation parameter. LLM-based verification is used only for objective answer extraction and equivalence checking; preference-based generation benchmarks use GPT-4.1 as a pairwise judge.}
  \begin{tabular}{@{}p{0.18\linewidth}p{0.25\linewidth}p{0.22\linewidth}p{0.24\linewidth}@{}}
    \toprule
    \textbf{Benchmark} & \textbf{Generation setting} & \textbf{Scoring/aggregation} & \textbf{Judge or checker} \\
    \midrule
    GPQA-Diamond & \begin{tabular}[t]{@{}l@{}}0-shot; \texttt{T=0.7}; $n=4$\\\texttt{top\_p=0.95}; \texttt{top\_k=20}\\\texttt{rep.=1.05}; max 32K tokens\end{tabular} & pass@1/accuracy estimate with $k=1$ & Qwen2.5-14B-Instruct verifier; \texttt{T=0.2} \\
    HMMT25 & \begin{tabular}[t]{@{}l@{}}\texttt{T=0.6}; $n=16$\\\texttt{top\_p=0.95}; \texttt{top\_k=20}\\\texttt{rep.=1.05}; max 32K tokens\end{tabular} & pass@1/accuracy estimate with $k=1$ & Qwen2.5-14B-Instruct verifier; \texttt{T=0.2} \\
    ZebraLogic & \begin{tabular}[t]{@{}l@{}}\texttt{T=0}; $n=1$\\max 32K tokens\end{tabular} & Puzzle/grid correctness & ZebraLogic rule-based grid evaluator \\
    LiveCodeBench v5 & \begin{tabular}[t]{@{}l@{}}0-shot; \texttt{T=0}; $n=1$\\max 32K tokens\end{tabular} & Code pass rate & Official execution-based evaluator \\
    LiveCodeBench v6 & \begin{tabular}[t]{@{}l@{}}0-shot; \texttt{T=0}; $n=1$\\max 32K tokens\end{tabular} & Code pass rate & Official execution-based evaluator \\
    IF-Eval & \begin{tabular}[t]{@{}l@{}}\texttt{T=0}; $n=1$\\max 16K tokens\end{tabular} & Instruction-following score & Official deterministic rule checker \\
    Arena-Hard v2 (HP) & \begin{tabular}[t]{@{}l@{}}\texttt{T=0}; $n=1$\\max 32K tokens\end{tabular} & Pairwise preference score & GPT-4.1 judge via Arena-Hard pipeline; \texttt{T=0.0} \\
    Arena-Hard v2 (CW) & \begin{tabular}[t]{@{}l@{}}\texttt{T=0}; $n=1$\\max 32K tokens\end{tabular} & Pairwise preference score & GPT-4.1 judge via Arena-Hard pipeline; \texttt{T=0.0} \\
    LiveBench 1125 & \begin{tabular}[t]{@{}l@{}}\texttt{T=0}; $n=1$\\max 32K tokens\end{tabular} & Official mixed benchmark score & Official exact, rule-based, symbolic, and execution scorers \\
    \bottomrule
  \end{tabular}
  \label{tab:evaluation_protocols}
\end{table*}

\section{Evaluation Protocols}
\label{sec:evaluation_appendix}

This appendix describes how the general-capability, role-play, and medical reasoning QA benchmarks are scored. For general capability, we use GPQA-Diamond~\citep{gpqa}, HMMT25 from MathArena~\citep{matharena}, ZebraLogic~\citep{zebralogic}, LiveCodeBench v5 and LiveCodeBench v6~\citep{livecodebench}, IF-Eval~\citep{ifeval}, the Hard Prompt and Creative Writing subsets of Arena-Hard v2~\citep{arena_hard}, and LiveBench 1125~\citep{livebench}. Table~\ref{tab:evaluation_protocols} summarizes the generation settings, aggregation rules, and judges or checkers used for these general benchmarks. We distinguish three evaluation modes. First, objective benchmarks with short canonical answers are scored by extracting the model's final answer and comparing it with the reference answer. Second, code and instruction-following benchmarks are evaluated by their official execution or rule-based checkers. Third, preference-based generation benchmarks use an LLM judge because the target quality criteria are preference-based rather than exact-answer based.

\paragraph{Answer Verification for Objective Benchmarks.}
For GPQA-Diamond and HMMT25, we use an LLM-based answer verifier implemented with Qwen2.5-14B-Instruct. The verifier is given the original prompt, the model response, and the ground-truth answer. It is instructed to identify the final answer in the response and output both an extracted answer and a binary decision indicating whether it matches the reference. Correctness is determined by this binary equivalence decision, with Qwen2.5-14B-Instruct serving as the answer extractor and equivalence verifier.

Concretely, the verifier prompt wraps each example as \texttt{<prompt>}, \texttt{<response>}, and \texttt{<ground\_truth\_answer>} fields and asks for \texttt{<extracted\_answer>} and \texttt{<decision>} tags. A prediction is counted as correct if and only if the parsed decision is \texttt{yes}. For multi-sample settings, we generate $n$ candidate responses and report the $k=1$ pass@1/accuracy estimate obtained from their binary decisions.

On a validation subset of 100 model outputs, with 50 outputs from each benchmark, the verifier decisions agreed with the official rule-based extraction for all examples, supporting its reliability for answer extraction and equivalence checking.

\paragraph{Rule-based and Execution-based Benchmarks.}
ZebraLogic is evaluated with the ZebraLogic grid evaluator, which parses the predicted grid and compares it against the puzzle solution. LiveCodeBench v5 and v6 are evaluated with the benchmark execution pipeline: generated programs are extracted and run against the associated tests, and pass rates are reported following the benchmark protocol. IF-Eval is scored with its official instruction-following checker, which applies deterministic constraints to the generated responses. LiveBench 1125 is evaluated with the official LiveBench ground-truth evaluation protocol, which applies the appropriate exact, rule-based, symbolic, or execution-based scorer to each task family.

\paragraph{Preference-based Generation Benchmarks.}
Arena-Hard v2 requires model-based evaluation because its outputs are judged by pairwise preference. We use GPT-4.1 as the judge and follow the Arena-Hard automatic judging pipeline. We report the Hard Prompt and Creative Writing subsets separately as HP and CW.

\subsection{Role-Play Evaluation Protocols}

For role-play preservation, we follow the official Given-Circumstance Acting (GCA) evaluation protocol from CoSER~\citep{coser}. The protocol simulates role-playing dialogues from the provided character profiles, scenario, and reference dialogue, and evaluates the generated dialogue using the CoSER scoring logic. In our experiments, Qwen2.5-72B-Instruct serves as the fixed environment model, next-speaker predictor, and judge, while the evaluated model varies. We also evaluate the models in Table~\ref{tab:main_results} using the official CoSER GPT-4o-based GCA configuration and obtain the same model ranking. Given the substantial cost of multi-turn dialogue simulation and judging, we use the Qwen2.5-72B-Instruct configuration for all reported main, ablation, and hyperparameter evaluations. We report the four standard GCA dimensions---Storyline Consistency, Anthropomorphism, Character Fidelity, and Storyline Quality---and use their arithmetic mean as the role-play average.

\subsection{Medical Reasoning QA Evaluation Protocols}

For the medical reasoning QA instantiation, we evaluate domain preservation with two objective benchmarks: MedXpertQA Text~\citep{medxpertqa} and MedQA-USMLE~\citep{medqa}.

\paragraph{MedXpertQA Text.}
MedXpertQA Text is a text-only expert-level medical multiple-choice benchmark from the Text subset of MedXpertQA~\citep{medxpertqa}\footnote{\nolinkurl{https://huggingface.co/datasets/TsinghuaC3I/MedXpertQA}}. We evaluate the full test split, which contains 2,450 examples. The model is prompted to choose a single option and end with an explicit final answer. We extract the predicted option letter from the final-answer pattern and compute exact-match accuracy against the reference label.

\paragraph{MedQA-USMLE.}
MedQA-USMLE~\citep{medqa} is a four-option medical exam benchmark. We use the \texttt{GBaker/MedQA-USMLE-4-options} test split, which contains 1,273 examples. The model is prompted as a multiple-choice question-answering system and is required to end with \texttt{Final answer: <letter>}. We extract the predicted option and compute exact-match accuracy.

\begin{table*}[t]
  \centering
  \small
  \renewcommand{\arraystretch}{1.12}
  \caption{Evaluation protocol for Medical Reasoning QA benchmarks. All scores are objective exact-match metrics.}
  \begin{tabular}{@{}p{0.22\textwidth}p{0.34\textwidth}p{0.36\textwidth}@{}}
    \toprule
    \textbf{Benchmark} & \textbf{Evaluation set} & \textbf{Scoring/aggregation} \\
    \midrule
    MedXpertQA Text & Full text test split; 2,450 examples & Multiple-choice exact match from parsed final-answer letter \\
    MedQA-USMLE & \texttt{GBaker/MedQA-USMLE-4-options}\newline test split; 1,273 examples & Four-option exact match from parsed final-answer letter \\
    \bottomrule
  \end{tabular}
  \label{tab:medical_evaluation_protocols}
\end{table*}


\section{Hyperparameter Analysis}
\label{sec:hyperparameter_appendix}

We further study the training schedule and two hyperparameters in the general recovery branch. The recovery-to-preservation schedule controls how often general recovery is updated relative to domain preservation. The recovery scaling ratio $r_g$ multiplies the gap signal of selected general-branch samples in Eq.~\eqref{eq:camopd_branch_loss}. The gap-score mass target $\rho_g$ controls how much total recovery gap-score mass must be covered by the selected samples in Eq.~\eqref{eq:mass_selection}. The main-text configuration uses a 3-1 recovery-to-preservation schedule, $r_g=2$, and $\rho_g=0.8$. We compare it with Vanilla MOPD and schedule, scaling, and mass-target variants, keeping the teachers, data, and other optimization settings unchanged.

\begin{table*}[t]
  \caption{General and Role-Play capability under different recovery schedules and general-recovery hyperparameters. Higher is better for all metrics.}
  \label{tab:hyperparams}
  \centering
  \scriptsize
  \setlength{\tabcolsep}{2pt}
  \resizebox{\textwidth}{!}{%
  \begin{tabular}{@{}lccccccccc@{}}
    \toprule
    \textbf{Metric}
    & \shortstack[c]{\textbf{Vanilla}\\\textbf{MOPD}}
    & \shortstack[c]{\textbf{1-1}\\$r_g{=}2$\\$\rho_g{=}0.8$}
    & \shortstack[c]{\textbf{5-1}\\$r_g{=}2$\\$\rho_g{=}0.8$}
    & \shortstack[c]{\textbf{3-1}\\$r_g{=}1$\\$\rho_g{=}0.6$}
    & \shortstack[c]{\textbf{3-1}\\$r_g{=}1$\\$\rho_g{=}0.8$}
    & \shortstack[c]{\textbf{3-1}\\$r_g{=}2$\\$\rho_g{=}0.6$}
    & \shortstack[c]{\textbf{3-1}\\$r_g{=}2$\\$\rho_g{=}0.8$}
    & \shortstack[c]{\textbf{3-1}\\$r_g{=}3$\\$\rho_g{=}0.6$}
    & \shortstack[c]{\textbf{3-1}\\$r_g{=}3$\\$\rho_g{=}0.8$}
    \\
    \midrule
    \multicolumn{10}{l}{\textit{General capability}} \\
    \midrule
    GPQA-Diamond & 61.99 & 61.11 & 60.23 & 61.87 & 61.62 & 61.62 & \textbf{62.63} & 60.98 & 62.50 \\
    ZebraLogic & 71.30 & 74.10 & \textbf{76.70} & 74.40 & 74.20 & 74.50 & 76.30 & 74.50 & 75.50 \\
    HMMT25 & 29.17 & 31.46 & 29.79 & 31.46 & 30.83 & 31.25 & 31.04 & 30.42 & \textbf{31.88} \\
    LiveCodeBench v5 & 31.54 & \textbf{37.99} & 37.28 & 34.41 & 35.13 & 34.77 & 37.28 & 34.41 & 35.48 \\
    LiveCodeBench v6 & 22.29 & 23.43 & 23.43 & \textbf{24.00} & 20.57 & \textbf{24.00} & \textbf{24.00} & 22.29 & 22.86 \\
    IF-Eval & 79.67 & 79.48 & 79.85 & 78.74 & 78.93 & 80.78 & 80.41 & 79.67 & \textbf{80.96} \\
    Arena-Hard v2 (HP) & 25.50 & 30.60 & \textbf{35.20} & 27.60 & 26.60 & 32.60 & 33.20 & 30.00 & 29.30 \\
    Arena-Hard v2 (CW) & 13.70 & 27.60 & 31.10 & 22.10 & 25.50 & 27.30 & \textbf{33.30} & 28.00 & 30.20 \\
    LiveBench 1125 & 58.90 & 63.00 & 63.20 & 59.70 & 61.40 & 62.70 & \textbf{63.50} & 63.00 & 62.30 \\
    \midrule
    \multicolumn{10}{l}{\textit{Role-Play domain capability}} \\
    \midrule
    Storyline Consistency & 43.57 & 45.29 & 45.30 & 46.79 & 44.46 & 46.14 & \textbf{47.81} & 44.58 & 47.47 \\
    Anthropomorphism & 34.91 & \textbf{36.58} & 35.98 & 34.24 & 35.50 & 34.34 & 33.96 & 35.65 & 34.05 \\
    Character Fidelity & 41.45 & 41.76 & 42.62 & 42.18 & 42.85 & 44.66 & 44.65 & \textbf{44.69} & 42.39 \\
    Storyline Quality & 51.50 & 52.10 & 51.55 & \textbf{54.58} & 53.22 & 53.08 & 53.47 & 53.88 & 54.25 \\
    Average & 42.86 & 43.93 & 43.86 & 44.45 & 44.01 & 44.56 & \textbf{45.00} & 44.70 & 44.54 \\
    \bottomrule
  \end{tabular}
  }
\end{table*}

\paragraph{Effect of Recovery-to-Preservation Schedule.}
Under $r_g=2$ and $\rho_g=0.8$, the 3-1 schedule gives the best recovery-preservation trade-off. The 5-1 schedule achieves similar general recovery, but its role-play average is lower than the 3-1 setting, indicating that domain capability starts to decline when recovery updates become too frequent. The 1-1 schedule is more conservative for recovery and also yields a lower domain average than 3-1.

\paragraph{Effect of Recovery Scaling and Gap-score Mass Target.}
For a fixed 3-1 schedule, $r_g=2$ and $\rho_g=0.8$ gives the strongest aggregate general recovery. The larger gap-score mass target covers a broader set of informative samples, while the recovery scale amplifies the gradient signal of the selected high-demand samples. Other combinations can improve individual metrics, but their gains are less consistent.

\paragraph{Trade-off Across General and Domain Performance.}
Based on these results, we use the 3-1 schedule with $r_g=2$ and $\rho_g=0.8$ in the main experiments. This configuration balances general recovery and role-play preservation across the evaluated benchmark families.


\section{Detailed Ablation Results}
\label{sec:ablation_appendix}

Tables~\ref{tab:ablation_roleplay_detailed} and~\ref{tab:ablation_medical_detailed} report the per-metric Role-Play and Medical Reasoning QA results, respectively, for the four matched variants summarized in Table~\ref{tab:ablation_delta}.

\makeatletter
\setlength{\@dblfptop}{0pt}
\setlength{\@dblfpsep}{12pt}
\setlength{\@dblfpbot}{0pt plus 1fil}
\makeatother

\begin{table*}[t]
  \caption{Detailed ablation results for the Role-Play dialogue instantiation. The four variants follow the matched ablation chain in Table~\ref{tab:ablation_delta}. Higher is better for all metrics, and bold marks the best result among the compared variants.}
  \label{tab:ablation_roleplay_detailed}
  \centering
  \begingroup
  
  \scriptsize
  \setlength{\tabcolsep}{2pt}
  \renewcommand{\arraystretch}{1.08}
  \resizebox{\textwidth}{!}{%
  \begin{tabular}{@{}l*{15}{c}@{}}
    \toprule
    \textbf{Variant}
    & \multicolumn{10}{c}{\textit{General capability}}
    & \multicolumn{5}{c}{\textit{Role-Play domain capability}} \\
    \cmidrule(lr){2-11}\cmidrule(lr){12-16}
    & \shortstack[c]{\textbf{GPQA}\\\textbf{Diamond}}
    & \shortstack[c]{\textbf{Zebra}\\\textbf{Logic}}
    & \shortstack[c]{\textbf{HMMT}\\\textbf{25}}
    & \textbf{LCB v5}
    & \textbf{LCB v6}
    & \shortstack[c]{\textbf{IF}\\\textbf{Eval}}
    & \shortstack[c]{\textbf{Arena}\\\textbf{HP}}
    & \shortstack[c]{\textbf{Arena}\\\textbf{CW}}
    & \shortstack[c]{\textbf{LiveBench}\\\textbf{1125}}
    & \shortstack[c]{\textbf{Gen.}\\\textbf{Avg.}}
    & \shortstack[c]{\textbf{Story}\\\textbf{Cons.}}
    & \textbf{Anthro.}
    & \shortstack[c]{\textbf{Char.}\\\textbf{Fid.}}
    & \shortstack[c]{\textbf{Story}\\\textbf{Qual.}}
    & \shortstack[c]{\textbf{Dom.}\\\textbf{Avg.}} \\
    \midrule
    \textbf{CaMOPD} & \textbf{62.63} & \textbf{76.30} & \textbf{31.04} & \textbf{37.28} & 24.00 & \textbf{80.41} & \textbf{33.20} & \textbf{33.30} & \textbf{63.50} & \textbf{49.07} & \textbf{47.81} & 33.96 & \textbf{44.65} & \textbf{53.47} & \textbf{45.00} \\
    \textbf{w/o selection} & 60.98 & 74.10 & 28.13 & 34.41 & 23.43 & 77.63 & \textbf{33.20} & 31.10 & 61.90 & 47.21 & 44.78 & 32.53 & 43.63 & 52.88 & 43.46 \\
    \textbf{w/o selection and scaling} & 61.74 & 74.70 & 27.29 & 35.48 & \textbf{24.57} & 78.00 & 30.20 & 25.50 & 61.10 & 46.51 & 44.75 & 31.92 & 41.22 & 52.50 & 42.60 \\
    \textbf{w/o selection, scaling, and decoupling} & 61.49 & 75.20 & \textbf{31.04} & 35.13 & 21.71 & 78.19 & 29.50 & 22.10 & 60.50 & 46.10 & 43.36 & \textbf{34.49} & 39.36 & 50.79 & 42.00 \\
    \bottomrule
  \end{tabular}%
  }
  \endgroup
\end{table*}

\begin{table*}[t]
  \caption{Detailed ablation results for the Medical Reasoning QA instantiation. The four variants follow the matched ablation chain in Table~\ref{tab:ablation_delta}. Higher is better for all metrics, and bold marks the best result among the compared variants.}
  \label{tab:ablation_medical_detailed}
  \centering
  \begingroup
  
  \scriptsize
  \setlength{\tabcolsep}{3pt}
  \renewcommand{\arraystretch}{1.08}
  \resizebox{\textwidth}{!}{%
  \begin{tabular}{@{}l*{13}{c}@{}}
    \toprule
    \textbf{Variant}
    & \multicolumn{10}{c}{\textit{General capability}}
    & \multicolumn{3}{c}{\textit{Medical domain capability}} \\
    \cmidrule(lr){2-11}\cmidrule(lr){12-14}
    & \shortstack[c]{\textbf{GPQA}\\\textbf{Diamond}}
    & \shortstack[c]{\textbf{Zebra}\\\textbf{Logic}}
    & \shortstack[c]{\textbf{HMMT}\\\textbf{25}}
    & \textbf{LCB v5}
    & \textbf{LCB v6}
    & \shortstack[c]{\textbf{IF}\\\textbf{Eval}}
    & \shortstack[c]{\textbf{Arena}\\\textbf{HP}}
    & \shortstack[c]{\textbf{Arena}\\\textbf{CW}}
    & \shortstack[c]{\textbf{LiveBench}\\\textbf{1125}}
    & \shortstack[c]{\textbf{Gen.}\\\textbf{Avg.}}
    & \shortstack[c]{\textbf{MedQA}\\\textbf{USMLE}}
    & \shortstack[c]{\textbf{MedXpertQA}\\\textbf{Text}}
    & \shortstack[c]{\textbf{Med.}\\\textbf{Avg.}} \\
    \midrule
    \textbf{CaMOPD} & \textbf{61.99} & 72.60 & \textbf{34.58} & \textbf{56.63} & \textbf{29.71} & \textbf{59.89} & \textbf{16.40} & \textbf{31.80} & \textbf{49.00} & \textbf{45.84} & \textbf{85.78} & 23.63 & \textbf{54.71} \\
    \textbf{w/o selection} & 60.61 & \textbf{73.10} & 31.88 & \textbf{56.63} & 28.57 & 58.04 & 15.90 & 31.10 & 48.60 & 44.94 & 85.62 & 23.59 & 54.61 \\
    \textbf{w/o selection and scaling} & 61.36 & 71.70 & 31.46 & 55.56 & 28.00 & 58.23 & 15.90 & 31.40 & 48.10 & 44.63 & 85.62 & 23.55 & 54.59 \\
    \textbf{w/o selection, scaling, and decoupling} & 58.84 & 69.90 & 29.79 & 54.84 & 28.00 & 58.04 & 12.10 & 29.00 & 48.20 & 43.19 & 85.39 & \textbf{23.71} & 54.55 \\
    \bottomrule
  \end{tabular}%
  }
  \endgroup
\end{table*}

\end{document}